\documentclass{article}
\usepackage{ijcai24}

\usepackage{times}
\usepackage{soul}
\usepackage{url}
\usepackage[hidelinks]{hyperref}
\usepackage[utf8]{inputenc}
\usepackage[small]{caption}
\usepackage{graphicx}
\usepackage{amsmath}
\usepackage{amsthm}
\usepackage{booktabs}
\usepackage{algorithm}
\usepackage{algorithmic}
\usepackage[switch]{lineno}

\usepackage{dsfont}
\usepackage{multirow}
\usepackage{svg} 
\usepackage{makecell}
\usepackage{epsfig}
\usepackage{amssymb}
\usepackage{subfig}

\title{When Fairness Meets Privacy: Exploring Privacy Threats in Fair Binary Classifiers through Membership Inference Attacks}

\author{
Huan Tian$^1$\and
Guangsheng Zhang$^1$\and
Bo Liu$^1$\and
Tianqing Zhu$^{2}$\and
Ming Ding$^3$\And
Wanlei Zhou$^2$\\
\affiliations
$^1$University of Technology Sydney\\
$^2$City University of Macau\\
$^3$Data61 CSIRO\\
}

\begin{document}

\maketitle

\begin{abstract}
While in-processing fairness approaches in machine learning show promise in mitigating bias predictions, 
their potential impact on privacy issues remains under-explored.
We aim to address this gap by assessing the privacy risks of fair enhanced binary classifiers with membership inference attacks (MIAs).
Surprisingly, 
our results reveal that these fairness interventions exhibit enhanced robustness against existing attacks, 
indicating that enhancing fairness may not necessarily lead to privacy compromises.
However, 
we find current attack methods are ineffective as they tend to degrade into simple threshold models with limited attack abilities.
Meanwhile, 
we identify a novel threat dubbed \textbf{F}airness \textbf{D}iscrepancy \textbf{M}embership \textbf{I}nference \textbf{A}ttacks (FD-MIA) that exploits prediction discrepancies between fair and biased models.
This attack demonstrates more effective attacks and poses real threats to model privacy. 
Extensive experiments across multiple datasets, attack methods, and representative fairness approaches validate our findings and demonstrate the efficacy of the proposed attack method.
Our study exposes the overlooked privacy threats in fairness studies, 
advocating for thorough evaluations of potential security vulnerabilities before model deployments.
\end{abstract}

\section{Introduction}
Imbalanced datasets often induce spurious correlations between learning targets and sensitive attributes~\cite{mehrabi2021survey}, which will lead to biased predictions in trained (\textit{biased}) models.
In response, 
fairness research has developed in-processing approaches~\cite{Wang_2022_CVPR,mroueh2021fair} that are applied during training to produce fairness-enhanced (\textit{fair}) models. 
By effectively suppressing these spurious correlations, 
fair models can mitigate discriminatory predictions.
However, 
despite promising to enhance fairness, 
these interventions might incur potential privacy risks, such as unintended training data memorization.

\begin{figure}[t]
    \centering
    \subfloat[Loss values\label{fig:intro_loss}]{
        \includegraphics[width=0.24\textwidth]{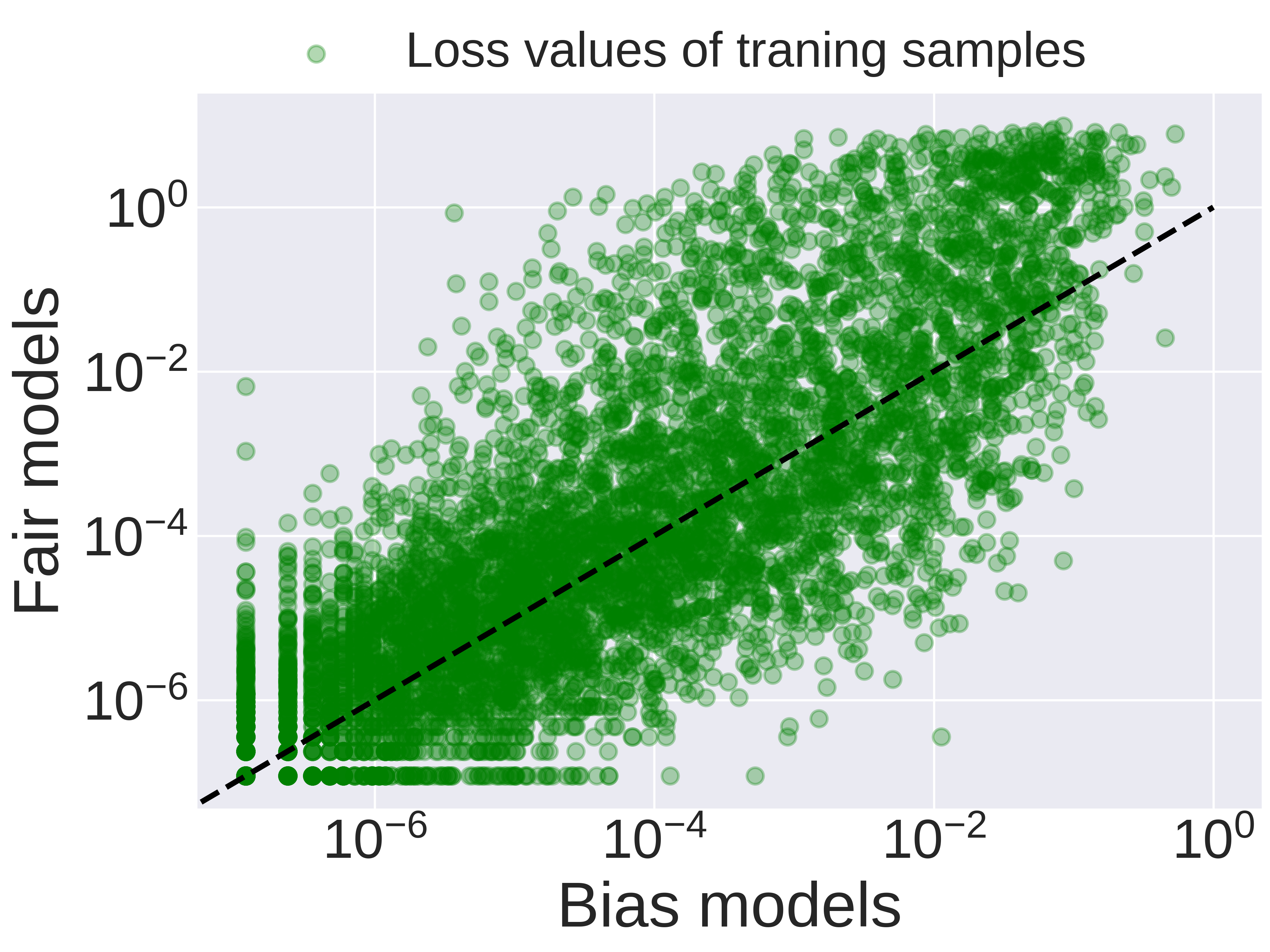}
    }
    \subfloat[Attack success rates\label{fig:intro_success}]{
        \includegraphics[width=0.24\textwidth]{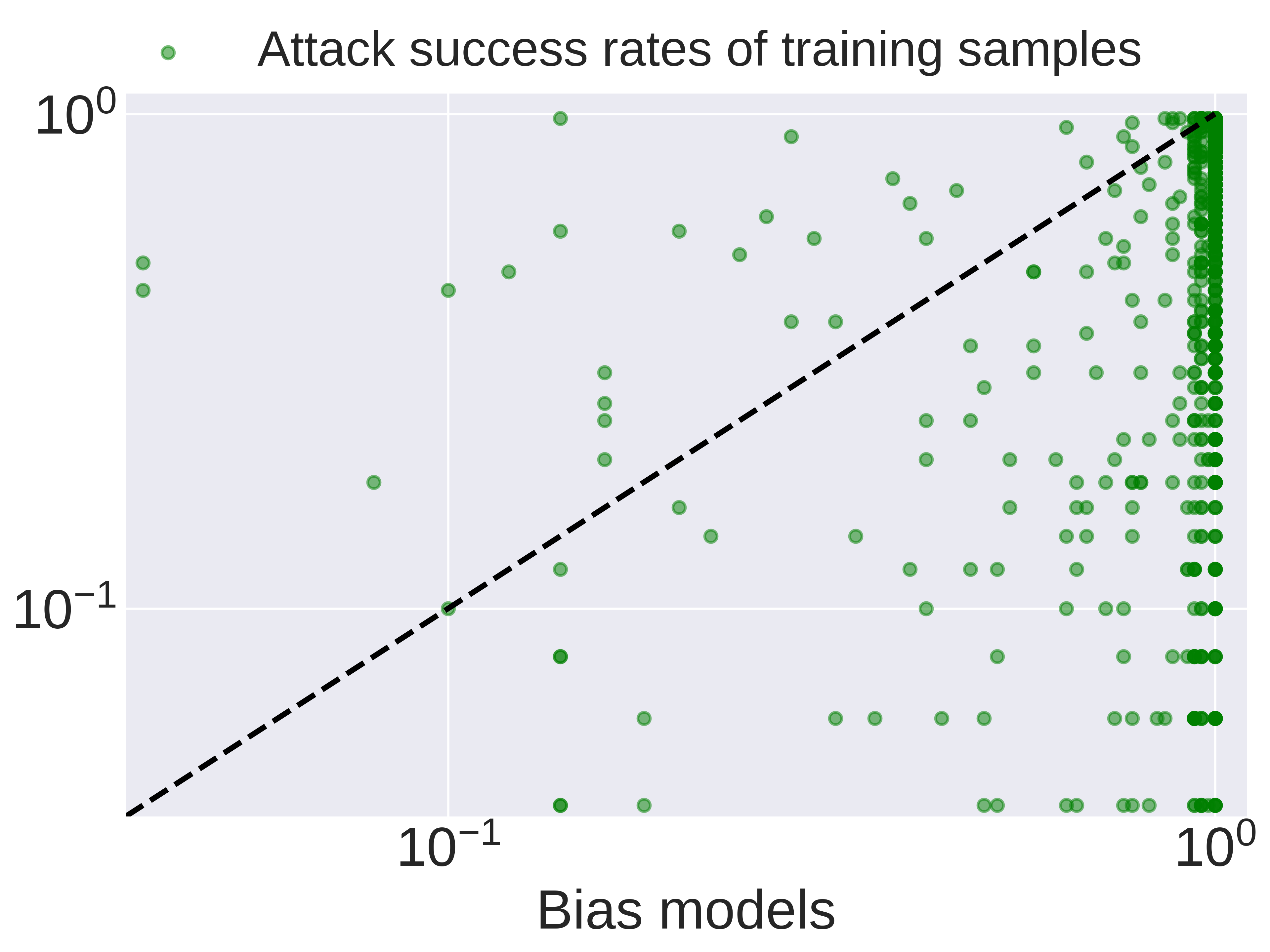}
    }
    \caption{Fairness interventions tend to (a) increase the losses and (b) decrease attack success rates for most training samples. We generate the plots with $100$ runs and report the mean loss value and mean attack success rate for the training sample.}
    \label{fig:intro}
\end{figure}

Membership inference attacks (MIAs) are widely adopted for assessing privacy risks in models deployed via Machine Learning as a Service (MLaaS)~\cite{shokri2017membership}.
Building upon this, 
prior work~\cite{chang2021privacy} has explored the privacy impact of fairness interventions by applying existing score-based attack methods to decision tree models.
However, 
this leaves open questions on evaluating privacy risks for neural networks, 
such as binary classifiers, which are prevalent in fairness studies.
Moreover, evaluations relying on one attack method might not fully capture the privacy impact.
Given the importance of trustworthiness in fairness studies, it is non-trivial to conduct thorough evaluations of these fairness interventions.
To bridge these gaps, we apply three different MIA methods to binary classifiers, including two existing MIA methods and one proposed method, comprehensively evaluating the privacy risk.

With existing attack methods, 
surprisingly, our evaluation results show that \textbf{fairness enhanced models are more robust to current MIAs than their biased counterparts.}
Figure~\ref{fig:intro} illustrates the results of score-based attack results on both fair and biased models over $100$ runs. 
We present the loss values (Figure~\ref{fig:intro_loss}) and report the attack success rate (Figure~\ref{fig:intro_success}) per training sample. 
The plots show increased loss values yet decreased attack success rates after applying fairness interventions for most data points. 
This indicates that these interventions can lead to less successful attacks with existing approaches.
However, 
we find that these methods are inefficient in attacking binary classifiers. 
This is because the trained attack models tend to degrade into simple threshold-based decisions due to the binary outputs.
The degradation incurs substantial performance trade-offs: while effective at recognizing member data, attack models struggle with non-member data, especially for hard samples where predictions are similar across groups. 
 
Besides the existing attack methods,
during the evaluation,
we have uncovered a potential threat that could enable more effective attacks on binary classifiers. 
Specifically, 
we find the prediction scores for member and non-member data tend to behave differently after fairness interventions: the scores tend to increase for member data, while they tend to align with a normal distribution for non-member data.
This creates a pronounced prediction gap between groups. 
The widened gap, 
if leveraged by adversaries, 
could enable more successful attacks, 
thereby posing substantial privacy threats.
Inspired by these observations, 
we name the identified threat as Fairness Discrepancy MIAs (FD-MIA), 
which exploits the prediction gaps between the original (biased) and fairness-enhanced (fair) models. 
FD-MIA can achieve superior attack performance by exploiting the prediction gaps.
It can also be flexibly combined with existing attack frameworks, 
such as score-based~\cite{LWHSZBCFZ22} and reference-based methods~\cite{carlini2022membership}.

We conduct experiments using six datasets, three attack methods, and five representative fairness approaches involving a total of $32$ settings and more than $160$ different models.
All results demonstrate our findings and the performance of the identified threat.
It reveals that \textbf{fairness interventions will introduce new threats to model privacy}, 
advocating for a more comprehensive examination of their potential security defects before deployment. Our main contributions are as follows:
(1) To the best of our knowledge, this is the first work to comprehensively study the impact of fairness interventions on privacy through the lens of MIAs, targeting deep learning classifiers. 
(2) We reveal that fairness interventions do not compromise model privacy with existing attack methods, primarily due to their limited efficiency in attacking binary classifiers.
(3) We identify and propose a novel attack method, FD-MIA, posing substantial threats to model privacy. 
It utilizes predictions from both biased and fair models and can be integrated into existing attack frameworks.
(4) Extensive experiments validate our findings, demonstrating the efficacy of FD-MIA.

\section{Related Work}
\noindent \textbf{Algorithmic fairness} methods are categorized into pre-processing, in-processing, and post-processing approaches based on their processing stage. We focus on in-processing approaches as they modify model training procedures directly and may pose model privacy threats. Among them, some introduce fair constraints and formulate the issues as optimization problems~\cite{Zemel:2013wz,Manisha2020FNNCAF,Tang_2023_CVPR,truong2023fredom,cruz2023fairgbm,jung2023reweighting}. Some propose adversarial designs to remove sensitive information among extracted features~\cite{Zhu_2021_ICCV,zafar2019fairness,Park2021LearningDR}.
Some adopt data sampling~\cite{NEURIPS2021_07563a3f} or reweighting~\cite{pmlr-v162-chai22a} approaches to alleviate the biased predictions.
More recently, 
studies learn fair representations using mixup augmentations or contrastive learning methods~\cite{mroueh2021fair,du2021fairness,Park_2022_CVPR,Wang_2022_CVPR,zhang2023fairness,qi2022fairvfl}. 
These methods interpolate inputs or modify features to pursue fair representations. 

\noindent \textbf{Membership inference attacks} aim to determine if a data sample was part of a target model's training set.
Some attacks leverage the target model's direct output, 
such as confidence scores~\cite{shokri2017membership}, losses~\cite{sablayrollesWhiteboxVsBlackbox2019}, 
prediction labels~\cite{choquette-chooLabelOnlyMembershipInference2021}.
Some improve the performance by modeling the prediction distributions of the target model, 
such as reference models~\cite{carlini2022membership,ye2022enhanced}. 
Others extend their focus into new settings~\cite{gaoSimilarityDistributionBased2023,yuanMembershipInferenceAttacks2022} or work on defense methods~\cite{yangPurifierDefendingData2023}.
This work considers two representative attack approaches: 
score-based~\cite{LWHSZBCFZ22} and reference-based~\cite{carlini2022membership} attack methods.
More recently, studies have enhanced attack performance by exploiting additional information:
some~\cite{heSemiLeakMembershipInference2022} leverage predictions from multiple augmented views, 
some~\cite{liAuditingMembershipLeakages2022} require results from multi-exit models, 
and others~\cite{hu2022m} work on multi-modality models. 
We explore models with fairness discrepancies. 

\noindent \textbf{Attacks on fairness} remain underexplored.
Prior study~\cite{chang2021privacy} provides an initial analysis by applying existing score-based MIAs to assess the privacy of fair constraint methods on decision tree models. They find fair decision trees enable more successful attacks.
In contrast,
we conduct comprehensive evaluations across diverse fairness approaches on neural network classifiers with multiple attack methods, including a potential threat that would compromise model privacy efficiently.

\section{Preliminaries}
\noindent \textbf{Algorithmic fairness.} 
Given biased models, 
we consider a sensitive attribute $s \in S$ with subgroups $\{s_0, s_1\}$ of binary attribute values $\{0,1\}$. Due to imbalanced training data, trained models often suffer from biased predictions, where predictions $ \widetilde{Y}$ become spuriously correlated with the attribute $s$. To mitigate this issue, various fairness methods have been proposed. To quantify model fairness, metrics such as \textit{Bias amplification} (BA)~\cite{zhao-etal-2017-men} or \textit{Equalized odds} (EO)~\cite{Hardt:2016wv} have been introduced. Specifically, BA measures disparities in true positives across subgroups, while EO measures both true and false positive rates (TPRs, FPRs). Our evaluations adopt BA and the Disparity of Equalized odds (DEO) to measure model fairness performance.

\noindent \textbf{Membership inference attacks.} \textit{Score-based attack methods} adopt the target model’s (i.e., models under attack) predictions (i.e., scores or losses) to infer sample membership.
To mimic target model behaviors, adversaries train ``shadow models'' on an auxiliary dataset that shares similar data distributions with the training data.
The outputs of shadow models can then be adapted for attack model training.
Formally,
given target models $\mathcal{T}$ and a queried sample $x \in X$, 
the membership $M(x)$ can be predicted by:
\begin{equation}
    M(x) = \mathds{1} [\mathcal{A}(\mathcal{T}(x))> \tau],
\label{mia_s}
\end{equation}
where the attack model $\mathcal{A}$ will output the confidence scores of membership predictions with a threshold $\tau$.

\noindent \textit{Reference-based likelihood ratio attack methods} 
infer the membership by modeling the prediction distributions. 
Specifically, they model the distributions using shadow models - $f_{\text{in}}$ that are trained with sample $x$, and $f_{\text{out}}$ that are trained without the $x$. The key idea is to determine if the target prediction $\mathcal{T}(x)$ better aligns with which of the prediction distributions.
Formally, membership is predicted by comparing the likelihood ratio $\Lambda$ between the two distributions:
\begin{equation}
   \Lambda=\frac{p(\phi(\mathcal{T}(x))|\mathcal{N}(\mu_{\text{in}}, \sigma_{\text{in}}))}{p(\phi(\mathcal{T}(x))|\mathcal{N}(\mu_{\text{out}}, \sigma_{\text{out}}))},
\label{lira}
\end{equation}
where $\phi$ is a scaling function, parameters $\mu$ and $\sigma$ are calculated with predictions from shadow models $f_{\text{in}}$, $f_{\text{out}}$. With the likelihood ratio, 
whichever is more likely determines the membership of $x$.

\section{Attacking Fair Models}
In this section, we first focus on a case study to explore the privacy impact of fairness interventions. We attack fair models using existing MIA methods, specifically the naive score-based attacks. 
We then introduce the proposed attack method that will pose substantial privacy threats.

\subsection{Naive score-based attacks}
\begin{figure}[t]
\begin{center}
\includegraphics[width=\linewidth]{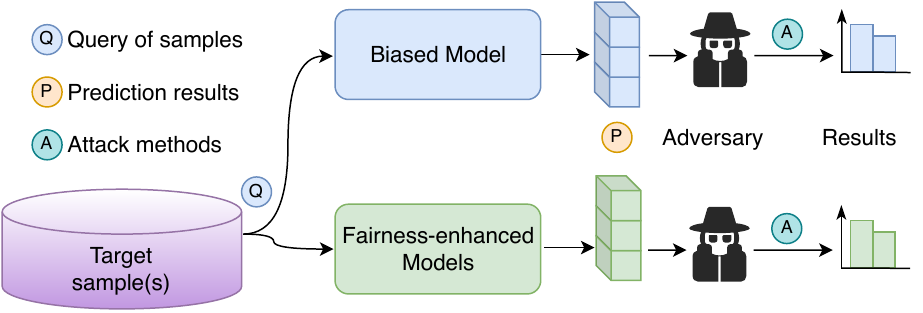}
\end{center}
\caption{Attacking fair and biased models with MIAs. We first attack them separately and then compare the results to explore the privacy impact of fairness interventions.}
\label{fig:pipeline}
\end{figure}

Figure~\ref{fig:pipeline} shows our evaluation pipeline with MIAs. 
We first train biased models with skewed data and then obtain fair models by applying fairness interventions. These models serve as target models for MIAs. An adversary attempts to infer sample membership with predictions from these target models. We then compare results across models to analyze the impact of fairness interventions on privacy.
In the following, we introduce detailed attack settings. 

\noindent \textbf{Target models.} We train biased models with the CelebA dataset \cite{CelebAMask-HQ}, which contains skewed data distributions for various attributes. In particular, we consider \textit{smile} as classification targets and \textit{gender} as the sensitive attribute. 
We train biased models following settings in \textit{ML-Doctor} from~\cite{LWHSZBCFZ22}.
We apply fair mixup operations from~\cite{mroueh2021fair,du2021fairness} to mitigate the biased predictions. 
Table \ref{table:score_results} presents accuracy ($\text{Acc}_\text{t}$) and fairness metrics ($\text{BA}$, $\text{DEO}$)
results for both biased (``Bias'') and fair (``Fair'') models.
The results show decreased fairness metric results, indicating the effectiveness of the adopted fairness interventions.

\noindent \textbf{Threat models.}
We apply naive score-based attacks on target models in a black-box manner. In particular, adversaries can only access models' predictions and an auxiliary dataset, which shares similar data distributions with the training data.
The adversary trains shadow models to mimic the target models' behavior and uses the prediction scores and results (\textit{true or false predictions}) to infer sample membership. 
In the experiments, we conduct attacks following settings in \textit{ML-Doctor} from~\cite{LWHSZBCFZ22}.

\noindent \textbf{Attack results.}
Table~\ref{table:score_results} shows the $\text{Acc}_\text{a}$ and $\text{AUC}_\text{a}$ results for attacks on the models. It shows improved robustness against MIAs after fairness interventions. For example, the accuracy results decreased from $59.8\%$ to $53.2\%$ with the fair models. AUC results exhibit similar trends. This aligns with results in Figure~\ref{fig:intro}, where fewer training samples can be successfully attacked after the interventions. Our initial results show that fairness interventions provide some defense against existing MIAs. However, our following analyses reveal that existing attacks are ineffective, especially for binary classifiers.

\begin{table}[t]
\centering 
  \begin{tabular}{c c c c c c}
  \toprule{Models}&
   {$\text{Acc}_\text{t}$ $\uparrow$} &
   {$\text{BA}$ $\downarrow$}&
   {$\text{DEO}$ $\downarrow$}&
   {$\text{Acc}_\text{a}$ $\uparrow$} &
   {$\text{AUC}_\text{a}$ $\uparrow$} 
    \\
    \midrule{Bias} &87.6 &7.7  &21.7  &59.8 &62.8  \\
            {Fair} &90.5 &2.5 &5.6 &53.2 &54.8  \\
    \bottomrule
  \end{tabular}
  \caption{Attack results with naive score-based methods in (\%).}
  \label{table:score_results}
\end{table}

\begin{figure}[t]
    \centering
    \subfloat[Attack accuracy \label{fig:flipper}]{
        \includegraphics[width=0.24\textwidth]{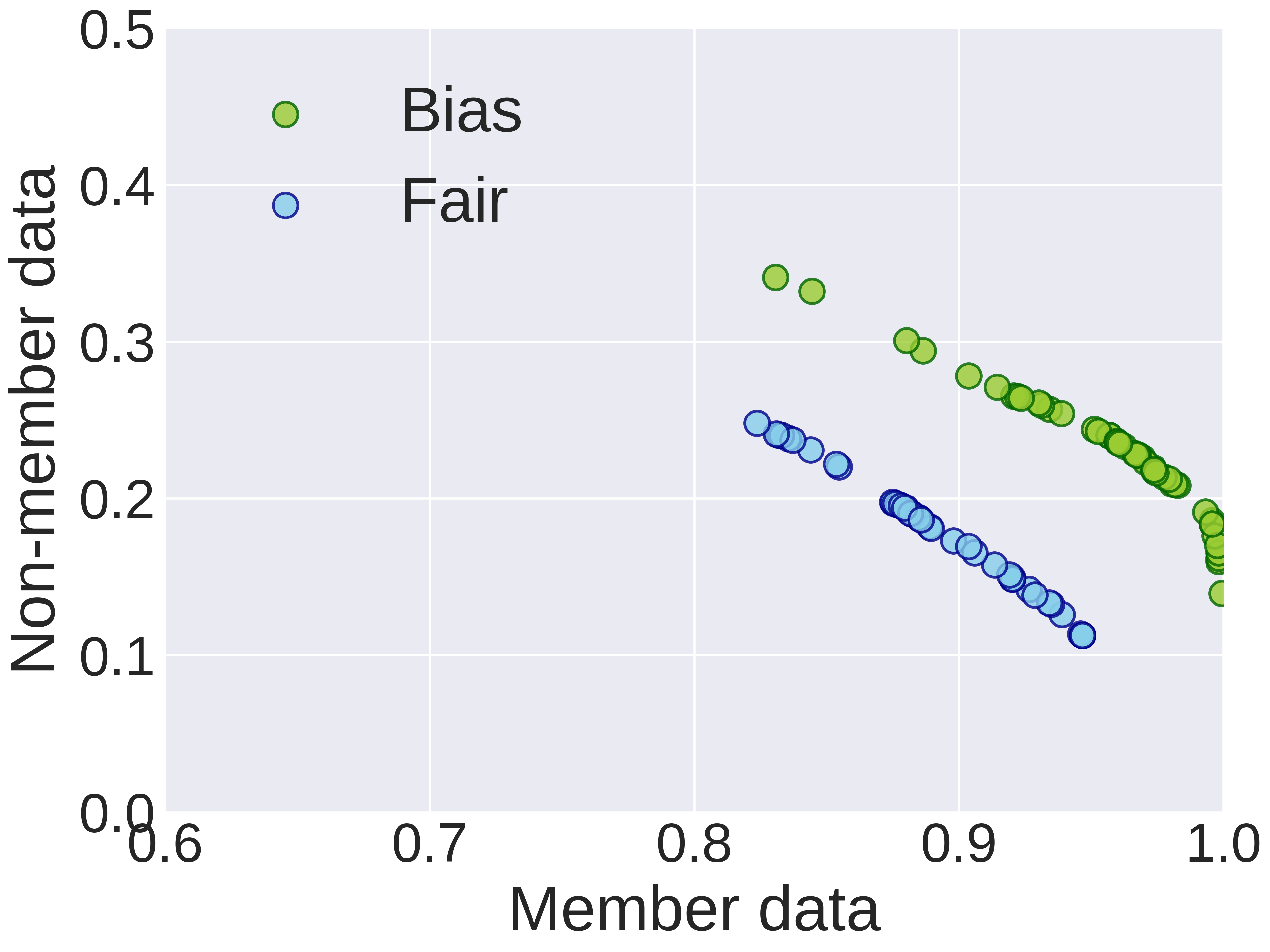}
    }
    \subfloat[Attack for hard examples \label{fig:score_fpr}]{
        \includegraphics[width=0.24\textwidth]{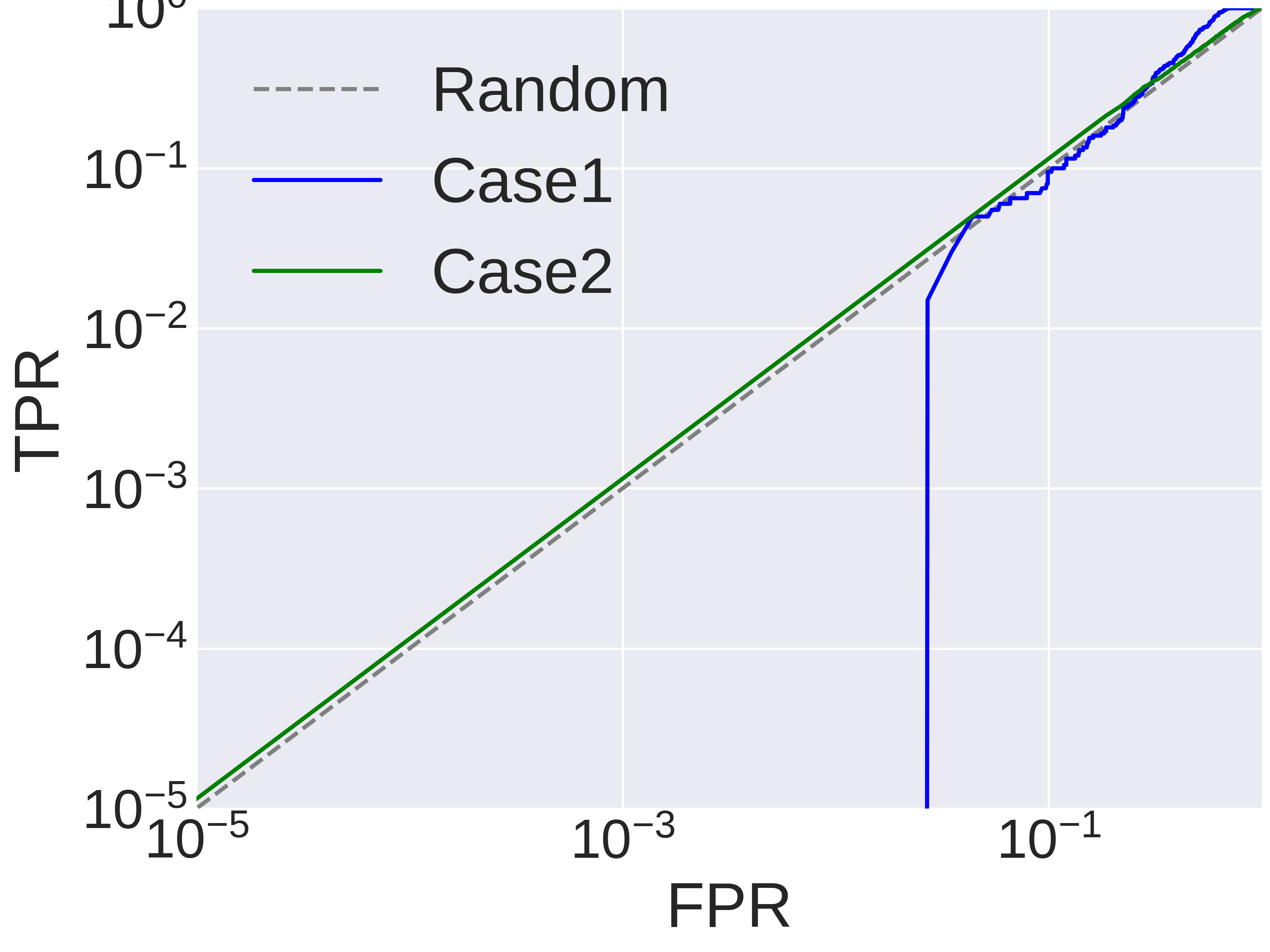}
    }
    \caption{Existing attacks (a) exhibit clear performance trade-offs between member and non-member data, and (b) are inefficient in attacking hard examples in the low FPR regime.}
    \label{fig:trade_off}
\end{figure}

\noindent \textbf{Performance trade-offs.} 
During the evaluation, 
we observe
\textbf{\textit{evident trade-offs in attack performance on member versus non-member data}} 
Figure~\ref{fig:flipper} depicts these trade-offs by comparing the accuracy results for member (x-axis) and non-member (y-axis) data.
We run over $100$ attacks on biased and fair models, and each point denotes one attack result.
The figure shows clear performance trade-offs between member and non-member data for both models. 
This raises concerns: whether achieving high attack performance comes at the cost of a higher false positive rate (FPR) on non-member data. 

The issue becomes more pronounced for hard examples where members and non-members share similar prediction scores. 
As suggested by~\cite{carlini2022membership}, we assess the attack performance for hard examples with TPR values in the low FPR regime.
We find the TPR values are around $0.0$ for most attacks. 
Figure~\ref{fig:score_fpr} presents two worst-case scenarios. The green curve in the figure shows closely aligned TPR and FPR values, indicating the attack results are equivalent to random guesses.
The blue line shows a TPR value of $0.0$ in low FPR regime, indicating that no positive samples can be correctly identified.
These findings reveal that attack models fail to differentiate the membership of hard examples,
revealing invalid attack method for hard examples.
This observation aligns with the concerns about the effectiveness of naive score-based attacks raised in previous studies~\cite{carlini2022membership,ye2022enhanced}. 

\begin{figure*}[t]
\begin{center}
\centering
    \subfloat[Biased models]{
        \label{fig:score_bias}
        \includegraphics[width=0.24\textwidth]{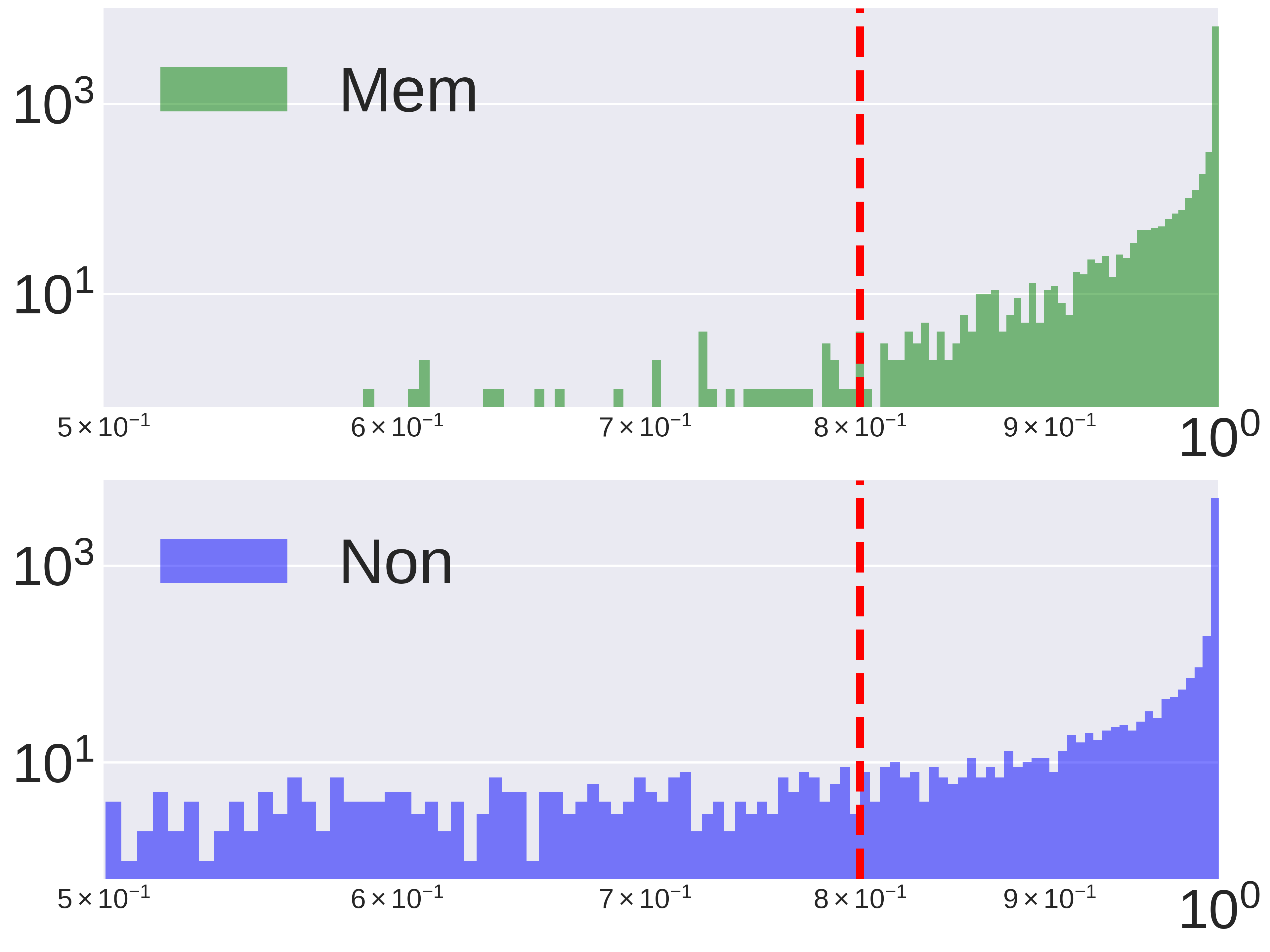}
    }
    \subfloat[Fair models]{\label{fig:score_fair}
        \includegraphics[width=0.24\textwidth]{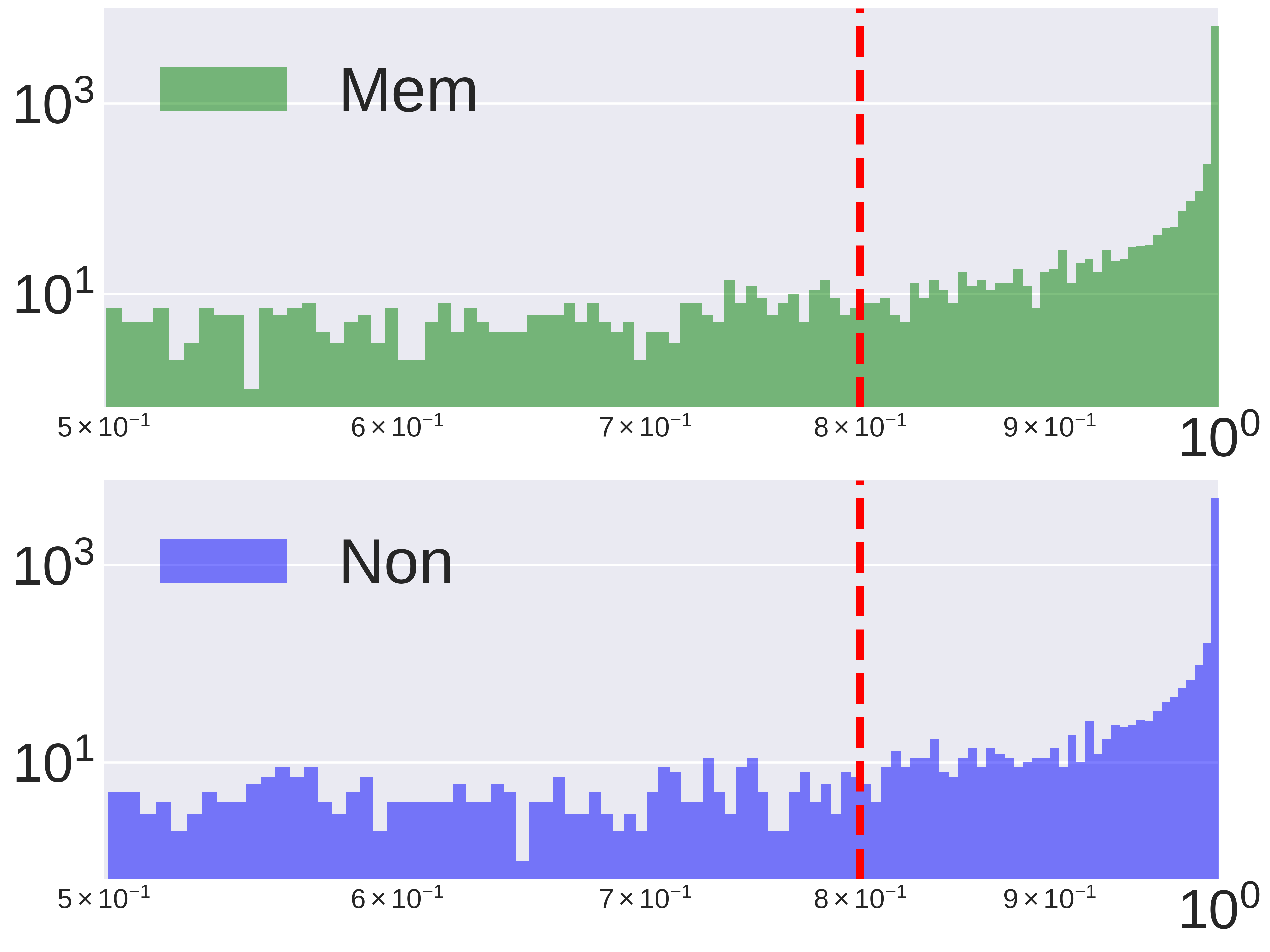}
    }
    \subfloat[Majority subgroups]{\label{fig:subg_majority}
        \includegraphics[width=0.24\textwidth]{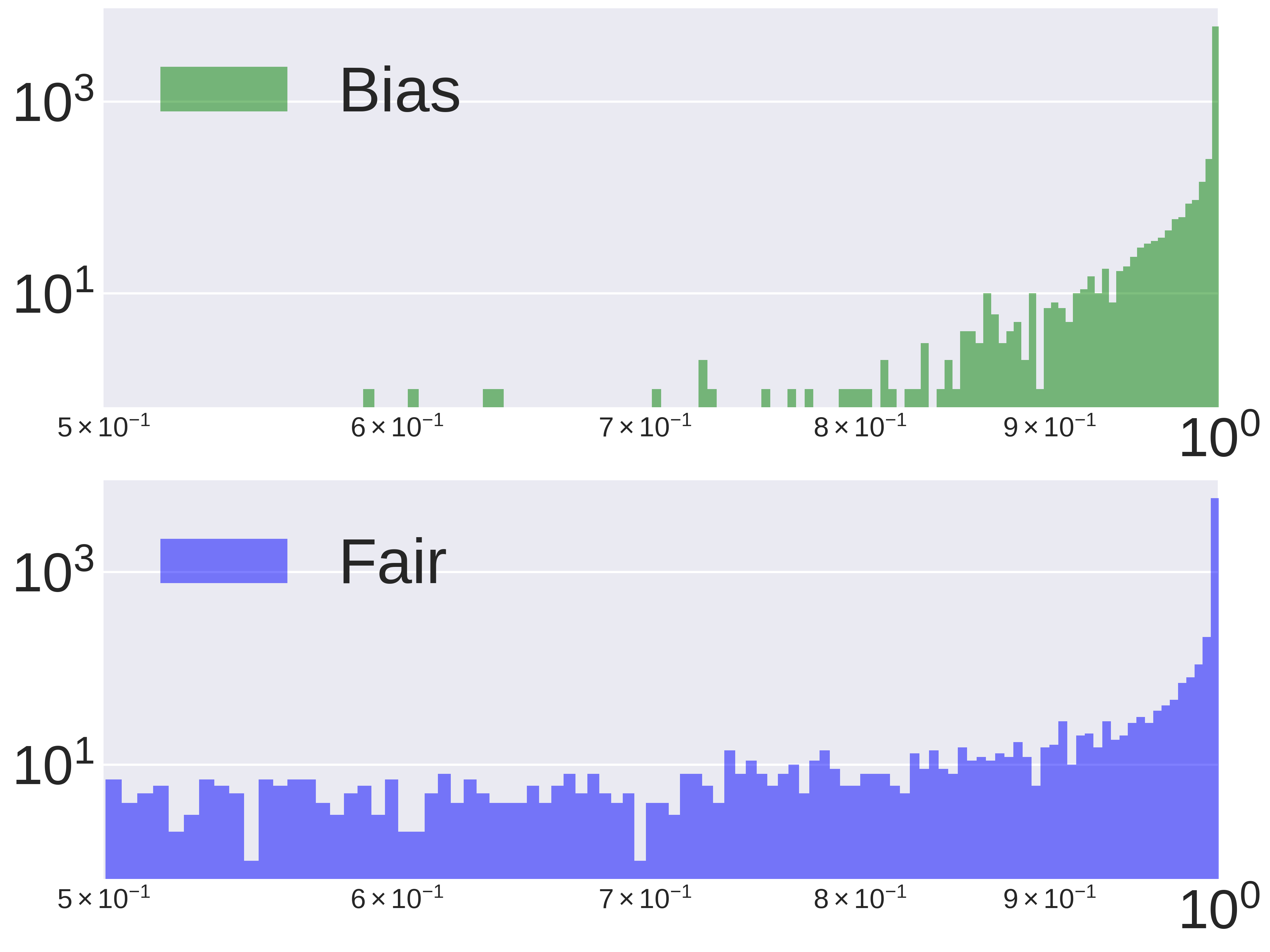}
    }
    \subfloat[Minority subgroups]{\label{fig:subg_minority}
        \includegraphics[width=0.24\textwidth]{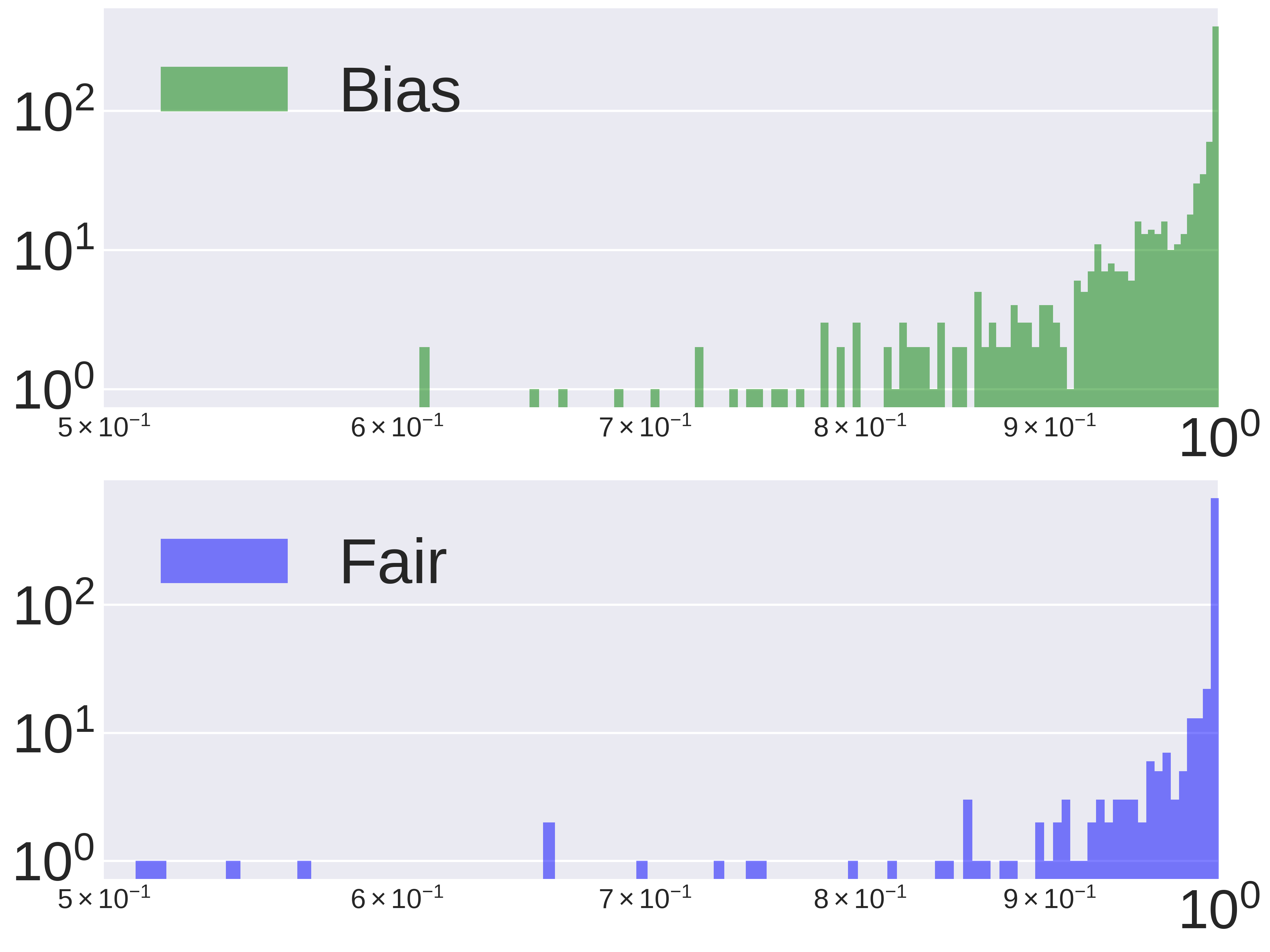}
    }
\end{center}
\caption{Prediction score changes after applying fairness methods. The red lines in (a) and (b) indicate that the trained attack models infer sample membership with certain threshold values. (c) and (d) show the changes in terms of different subgroups.}
\label{fig:conf_score}
\end{figure*}

\noindent \textbf{Model degradation.} 
To explore the reason, we further find that \textbf{\textit{trained attack models tend to degrade into simple threshold models with one-dimensional inputs.}} 
Current attack methods rely on prediction outcomes to determine the sample membership. 
However, 
in the context of binary classifiers, 
prediction scores can be reduced to one dimension as the sum of the confidence scores always equals one. 
Consequently, 
the attack model can essentially be viewed as a simple threshold model, 
which infers the membership by ``thresholding'' one-dimensional values. 

Figure~\ref{fig:score_bias} presents histograms of prediction scores with vertical lines representing some threshold values. 
By adjusting the vertical line (threshold), 
it is possible to achieve higher accuracy for member data, 
but this comes at the expense of decreased accuracy for non-member data. 
This threshold adjustment explains the aforementioned trade-off phenomenon.

\noindent \textbf{Impacts of fairness interventions.} 
When examining the prediction scores, 
we find that \textbf{\textit{fairness interventions decrease confidence scores for the majority training data, introducing some defense against existing MIAs.}}
This can be evidenced by the histograms of confident scores in Figures~\ref{fig:score_bias} and~\ref{fig:score_fair}.
The figures show that fairness interventions result in more similar score distributions between member and non-member data, making it challenging to distinguish with threshold attack models.

Moreover, 
we explore the score changes for different subgroups in Figures~\ref{fig:subg_majority} and~\ref{fig:subg_minority}. 
From the plots, the majority data tend to be more ``spread out'', whereas the minority are ``more concentrated''.  
This is because fairness interventions strive to balance prediction performance across subgroups for fair predictions.
The results advocate the observed increased loss values for most data points in Figure~\ref{fig:intro_loss} 
This observation also aligns with the fairness-utility trade-off, extensively observed in prior studies~\cite{zhang2023fairness,PinzonPPV22,Zietlow_2022_CVPR}, where achieving fairness often compromises model prediction performance.

Our analyses indicate that existing attack methods are ineffective in exploiting prediction gaps that could lead to model privacy leaks. While fairness interventions do introduce some defense to MIAs, we identify a novel threat that will impose substantial risks to model privacy.

\subsection{Attacks with FD-MIA}
\begin{figure}[t]
    \centering
    \subfloat[All data\label{fig:intro_general}]{
        \includegraphics[width=0.24\textwidth]{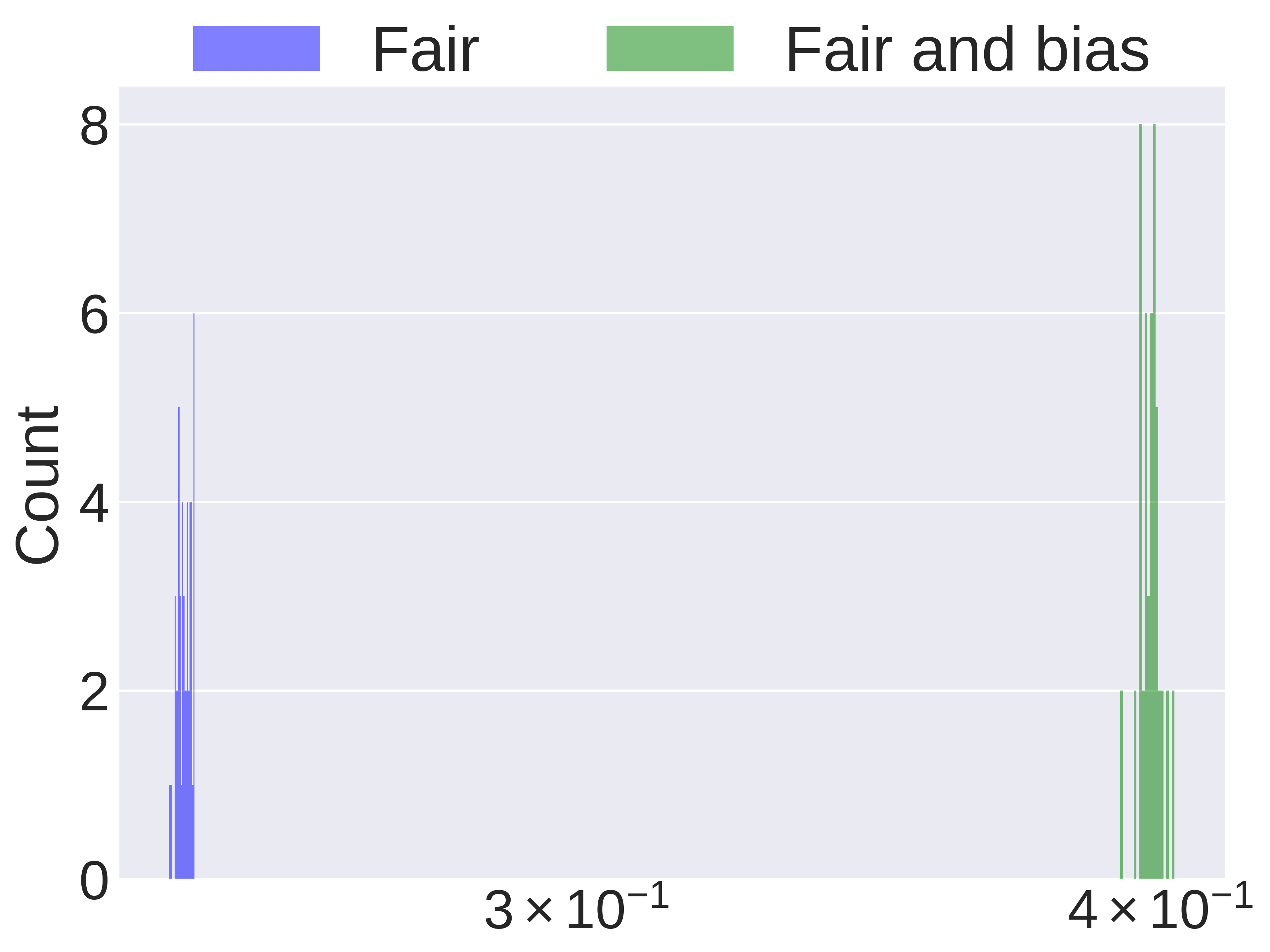}
    }
    \subfloat[Hard examples\label{fig:intro_hard}]{
        \includegraphics[width=0.24\textwidth]{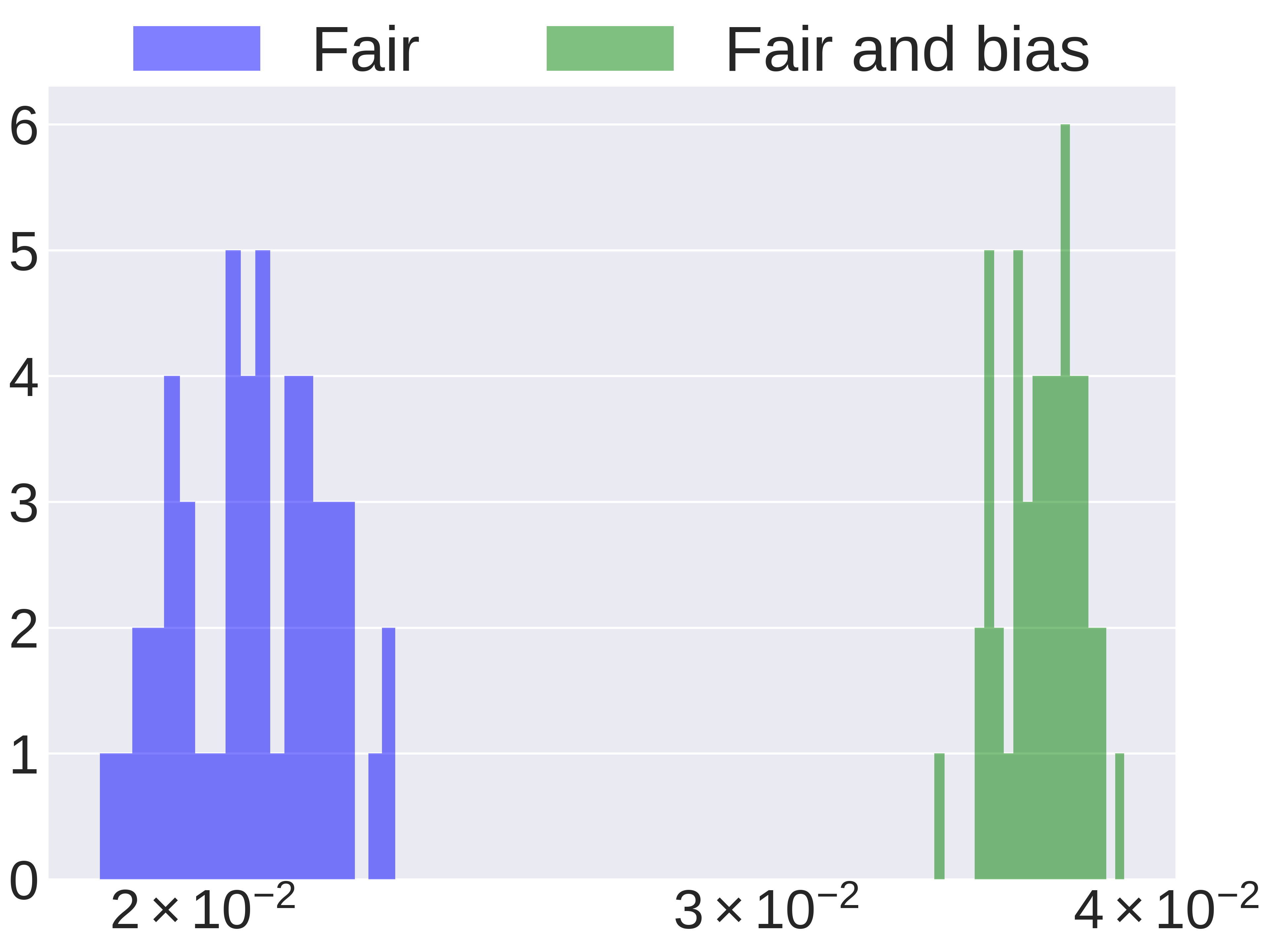}
    }
    \caption{Histograms of prediction score distances between member and non-member data. The plots show enlarged distance when considering both fair and biased models.}
    \label{fig:enlarged}
\end{figure}

\noindent \textbf{Enlarged prediction gaps.} Previous plots in Figures~\ref{fig:score_bias} and~\ref{fig:score_fair} show that fairness interventions reduce confidence scores for most training data. In contrast, non-member scores do not exhibit significant changes.
This disparity can increase prediction gaps between member and non-member data. 
We measure the gaps by calculating the score value differences between them and present the results in Figure~\ref{fig:enlarged}. 
To comprehensively evaluate the gaps, We measure the gaps with all available data (Figure~\ref{fig:intro_general}) and with only hard examples (Figure~\ref{fig:intro_hard}), where samples from the members and non-members share similar scores. 
The plots demonstrate a significant increase in distance when considering predictions from both biased and fair models. Importantly, the fairness interventions amplify these gaps, imposing real threats to model privacy.

\begin{figure}[t]
\begin{center}
\includegraphics[width=\linewidth]{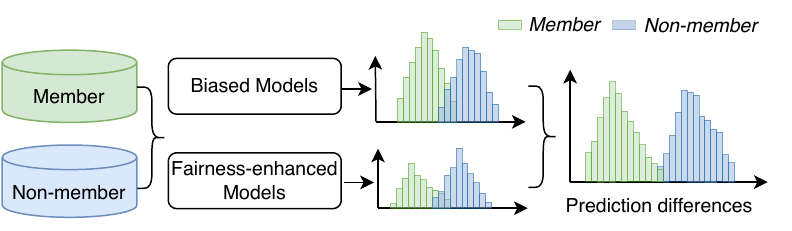}
\end{center}
\caption{FD-MIA exploits the predictions from both models to achieve efficient attacks.}
\label{fig:dual_flow}
\end{figure}

\noindent \textbf{Attack pipeline.} With the enlarged prediction gaps, we propose an enhanced attack method. We present the attack pipeline in Figure~\ref{fig:dual_flow}, where an adversary can access prediction results from both fair and biased models. The attack models will exploit the observed prediction gaps to infer sample membership. We refer to the proposed method as the \textit{Fairness Discrepancy based Membership Inference Attack} (\textit{FD-MIA}). 

\noindent \textbf{Threat models.} FD-MIA operates as a black-box attack and only needs access to predictions from both biased and fair models. In practice, adversaries could obtain such predictions, as real-world models often exhibit persistent biased predictions. For instance, they may monitor an MLaaS over time since debiasing should be continually carried out to adhere to legislation. Alternatively, they could deliberately report biases, compelling the owner to refine the model per regulations.
By recording the prediction shifts during these debiasing efforts, the adversary can enable efficient attacks.
Meanwhile, as FD-MIA only modifies the inputs, 
it can be seamlessly integrated into existing attack frameworks such as score-based and reference-based attacks.

\noindent \textbf{Score-based FD-MIA} introduces extra encoding layers to naive score-based attacks for extracting features of model predictions. Formally, compared to the naive score-based attacks in Eq.(\ref{mia_s}), the proposed method can be expressed as:
\begin{equation}
    M(x) = \mathds{1} [\mathcal{A}(\mathcal{T}_{\text{bias}}(x),\mathcal{T}_{\text{fair}}(x))> \tau],
\end{equation}
where the the attack models $\mathcal{A}$ takes model predictions of biased models $\mathcal{T}_{\text{bias}}$ and fair models $\mathcal{T}_{\text{fair}}$.

\noindent \textbf{Reference-based FD-MIA} incorporates with LiRA~\cite{carlini2022membership} and infer sample membership by modeling the prediction distributions. It enhances attack performance using two target models - the biased and the fair ones. Formally, for queried samples $x$ with target models $\mathcal{T}$, the membership probability is:
\begin{equation}
    p=(\phi(\mathcal{T}(x))|\mathcal{N}(\mu_{\text{bias}}, \mu_{\text{fair}}, \text{Cov})),
\end{equation}
where $\text{Cov}$ is the covariance matrix. Distribution function $\mathcal{N}$ takes the mean confidence scores of the biased $\mu_{\text{bias}}$ and fair models $\mu_{\text{fair}}$. The equation estimates the probability of members and non-members. The higher probability determines the inferred membership.

The proposed FD-MIA can lead to enhanced attack results as it leverages additional information for the attacks, which prevents the degradation of trained attack models. It demonstrates that fairness interventions can inadvertently introduce privacy risks that expose vulnerabilities to MIAs. 


\begin{table*}[t]
\centering 
\small
  \begin{tabular}{ c  c c c c c  c c c c c  c c c c c}
  \toprule \multirowcell{2}{Models}&
  \multicolumn{5}{c}{CelebA (T=s/S=g)} & \multicolumn{5}{c}{UTKFace (T=r/S=g)} &
  \multicolumn{5}{c}{FairFace (T=r/S=g)} \\
  \cmidrule{2-16}&
  {$\text{Acc}_\text{t}$} & {DEO} & {$\text{Acc}_\text{a}$} & {$\text{AUC}_\text{a}$} & {TPR} &
  {$\text{Acc}_\text{t}$} & {DEO} & {$\text{Acc}_\text{a}$} & {$\text{AUC}_\text{a}$} & {TPR} &
  {$\text{Acc}_\text{t}$} & {DEO} & {$\text{Acc}_\text{a}$} & {$\text{AUC}_\text{a}$} & {TPR} 
   \\
   \midrule
   {$\text{Bias}_{s}$} &87.6&21.7&59.8 &62.8  &0.0  &87.4&14.2& 58.5& 58.9& 0.0&  87.2&22.2&63.6 & 66.4& 0.0\\
   {$\text{Fair}_{s}$} &90.5&5.6&53.2 &54.8  &0.04  &89.0&6.3&52.6 & 52.8&0.0  &87.6&3.9& 63.3& 66.2& 0.0 \\
   {$\text{Our}_{s}$} &-&-&\textbf{60.6} &\textbf{65.8}  &\textbf{0.3}  &-&-& \textbf{60.2}& \textbf{62.1}& \textbf{0.2} &-&-&\textbf{65.2} &\textbf{66.8} &\textbf{0.2}\\
   \midrule
   {$\text{Bias}_{l}$} &87.6&21.7&51.5 &51.4  &0.6  &87.4&14.2&55.4 & 51.5& 0.9 &87.2&22.2 60.2 & 61.7& 1.3\\
   {$\text{Fair}_{l}$} &90.5&5.6&50.8 &50.3  &0.2 &89.0&6.3&53.2 & 47.6& 0.7 &87.6&3.9&56.7 & 57.2&0.9\\
   {$\text{Our}_{l}$} &-&-&\textbf{54.7} &\textbf{57.3}  &\textbf{1.2}  &-&-&\textbf{55.9} & \textbf{52.2}& \textbf{1.7} &-&-&\textbf{62.3} &\textbf{63.2} &\textbf{2.3}\\
   \bottomrule
  \end{tabular}
  \caption{Attacks for the \textit{gender} attribute (S), and different learning targets (T) in (\%), we report TPR@FPR of $0.1\%$.}
  \label{table:dual_results}
\end{table*}

\section{Experiments}
In this part, we extensively evaluate our findings and the proposed methods with multiple datasets. We first focus on the sensitive attribute of \textit{gender}. We then explore the results with other attributes and diverse ablation studies. We start by introducing the experiment settings.

\noindent \textbf{Settings.} With the \textit{gender} attribute, we consider following binary classifications: smiling predictions (T=s/S=g) with the CelebA dataset~\cite{CelebAMask-HQ}, race predictions (T=r/S=g) with the UTKFace dataset~\cite{geraldsutkface} and the FairFace dataset~\cite{karkkainen2021fairface}. As UTKFace and FairFace contain multiple racial subgroups, we first group them into \textit{White} and \textit{Others} and then obtain the binary subgroups.
For training data, 
we randomly divide the dataset in half to construct member and non-member data for MIAs. 
For target models, We sample the data with skewed data distributions considering the sensitive attribute to reflect real-world imbalances. 
This incurs biased predictions of trained models, which serve as the biased models.
We then apply fairness interventions of the mixup operations to obtain the fair models. 
We present more detailed experiment settings in \textit{Appendix A}.

\begin{figure}[t]
    \centering
    \subfloat[Score-based attacks \label{fig:lowFPR_score}]{
        \includegraphics[width=0.24\textwidth]{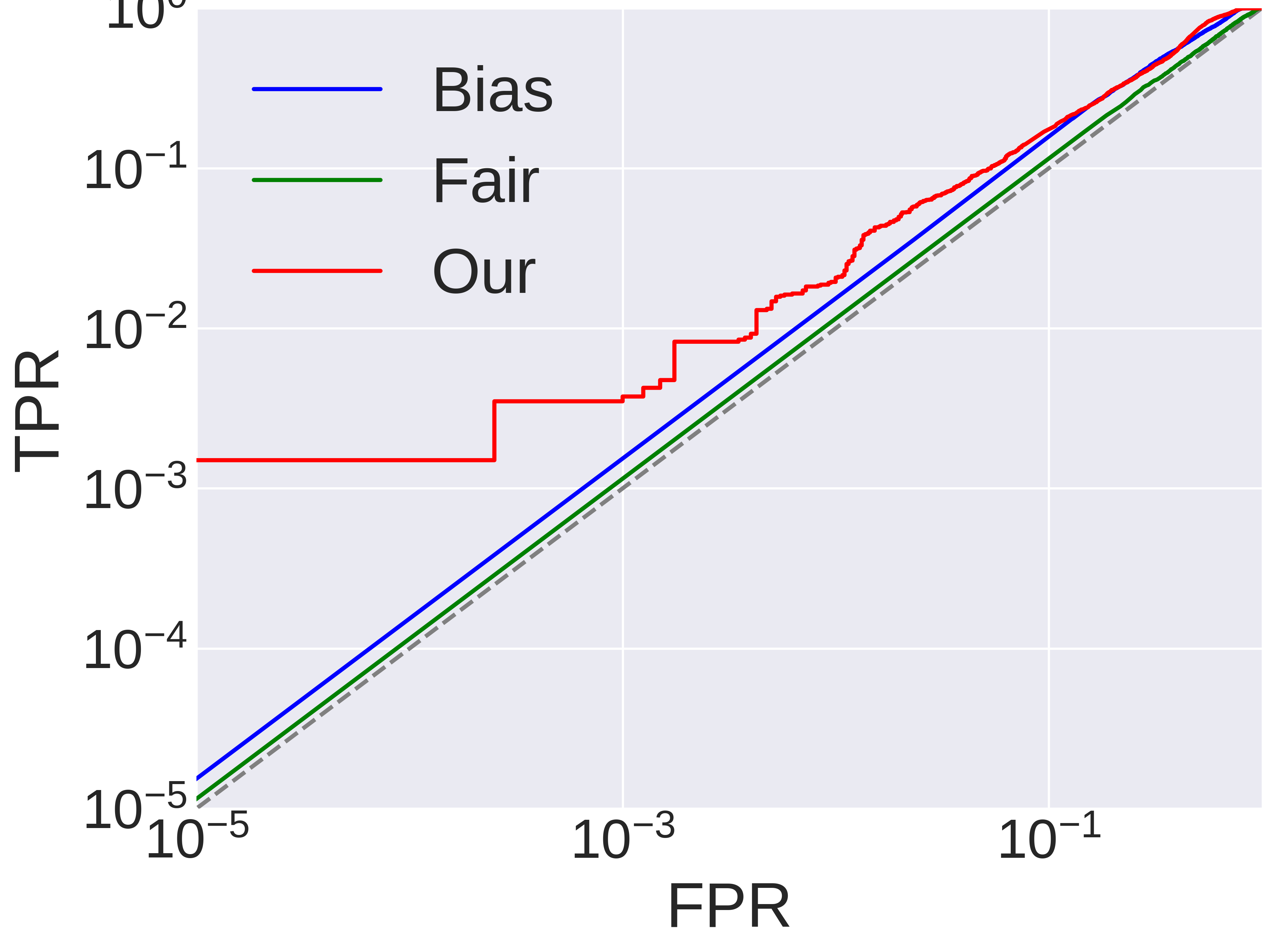}
    }
    \subfloat[LiRA attacks \label{fig:lowFPR_lira}]{
        \includegraphics[width=0.24\textwidth]{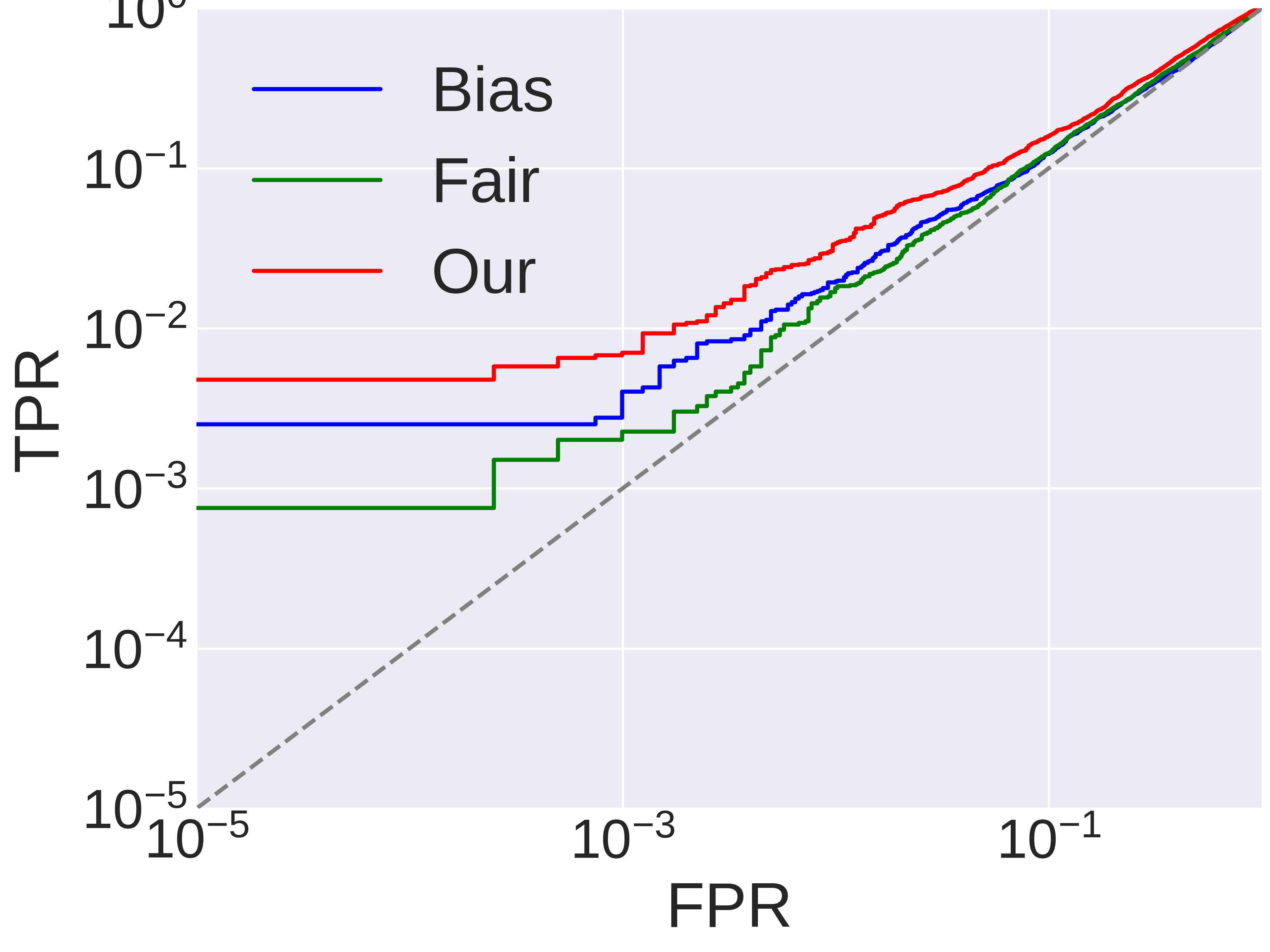}
    }
    \caption{Attack result comparisons in the low FPR regime for (a) score-based attacks and (b) LiRA attacks.}
    \label{fig:dual_fpr}
\end{figure}

\noindent \textbf{Results with the \textit{gender} attribute.} 
Table~\ref{table:dual_results} presents the attack results with different attack methods and metrics. 
We integrate the proposed FD-MIA with score-based attacks ($s$) and LiRA attacks ($l$). 
Besides the attack accuracy,
we further report the TPR results at a low FPR value of $0.1\%$, following suggestions in~\cite{carlini2022membership}. 
The table shows that FD-MIA outperforms all existing MIA methods with all cases and metrics. Notably, it achieves higher attack success on fair models than the biased ones. In contrast, the existing methods perform worse on fair models. This reveals that the proposed FD-MIA can effectively exploit fairness interventions to improve MIAs, posing threats to model privacy.

More specifically, in score-based attacks, FD-MIA outperformed others with the highest accuracy and AUC. Regarding TPR outcomes, notably, the existing attacks ($\text{Bias}_{s}$ and $\text{Fair}_{s}$) attain near-zero results. This indicates no true positive samples can be identified. In contrast, FD-MIA displayed valid TPR values, indicating effective attacks. Similar trends can be observed with the LiRA attacks. We further present the ROC curves for the CelebA case in Figure~\ref{fig:dual_fpr}. The figure further confirms the invalid attacks of the existing methods and the valid TPR results of FD-MIA.

Moreover, from the table, we observe that the attacks achieved superior results on FairFace compared to other datasets. Meanwhile, FairFace exhibits a greater discrepancy in fairness between biased and fair models. We believe the enlarged discrepancy leads to enlarged prediction gaps, enabling more effective attacks. Additionally, we notice that score-based attacks perform better on accuracy and AUC, whereas LiRA achieves better TPR values. This aligns with the observations in~\cite{carlini2022membership} as LiRA is designed for efficient attacks at the low FPR. We further present more detailed results in \textit{Appendix B}.

\begin{table*}[t]
\centering 
  \begin{tabular}{ c c c c  c c c  c c c c c c}
  \toprule \multirowcell{2}{Models}&
  \multicolumn{3}{c}{CelebA (T=s/S=h)} & \multicolumn{3}{c}{CelebA (T=s/S=m)} &
  \multicolumn{3}{c}{UTKFace (T=g/S=r)} &\multicolumn{3}{c}{FairFace (T=g/S=r)}\\
  \cmidrule{2-13}
   & {Acc} & {AUC} & {TPR@FPR} 
   & {Acc} & {AUC} & {TPR@FPR} 
   & {Acc} & {AUC} & {TPR@FPR}
   & {Acc} & {AUC} & {TPR@FPR}
   \\
   \midrule
   {$\text{Bias}_{s}$} &55.1 &56.3  &0.1 & 57.4&58.1 & 0.0&64.0 & 66.9& 0.0 &75.5&76.7& 0.0 \\
   {$\text{Fair}_{s}$} &52.6 &52.7  &0.0 & 53.1& 52.0&0.0 & 55.3& 57.2& 0.0&73.2 &75.5& 0.0 \\
   {$\text{Our}_{s}$} &\textbf{56.9} &\textbf{59.6}  &\textbf{0.2}  & \textbf{59.6}& \textbf{63.2}&\textbf{0.2}&\textbf{66.7} & \textbf{67.8}&\textbf{0.3} &\textbf{77.0} &\textbf{78.4}& \textbf{0.7}\\
   \midrule
   {$\text{Bias}_{l}$} &52.1 &52.0  &0.3  & 51.6& 51.4& 0.4&55.5 & 52.4 &1.4 &73.2&74.2&1.5\\
   {$\text{Fair}_{l}$} &51.0 &50.5  &0.1 &50.7 & 49.9& 0.1&53.8 &49.7 &0.9 &70.4&72.1&0.6\\
   {$\text{Our}_{l}$} &\textbf{55.4} &\textbf{57.7}  &\textbf{0.8}  & \textbf{54.2}& \textbf{55.7}& \textbf{0.6}&\textbf{56.2} & \textbf{53.6}& \textbf{2.1}&\textbf{75.2} &\textbf{76.4}&\textbf{2.9}\\
   \bottomrule
  \end{tabular}
  \caption{Attacks with different sensitive attributes and learning targets in (\%).}
  \label{table:other_attrs}
\end{table*}

\noindent \textbf{Results with other attributes.} We further explore attacks with different attributes, including \textit{wavy hair} (T=s/S=h) and \textit{heavy makeup} (T=s/S=m) for CelebA, as well as \textit{race} (T=g/S=r) for UTKFace and FairFace. Table~\ref{table:other_attrs} presents the results. Once again, the proposed FD-MIA outperforms existing methods on all datasets and metrics, posing real privacy threats. Notably, it consistently achieves superior performance with varying accuracy, ranging from $53\%$ to $73\%$. The results illustrate the robustness of FD-MIA, highlighting its efficacy in real-world scenarios. Additionally, similar to previous results, FD-MIA achieves better attack performance on FairFace, likely due to the enlarged fairness discrepancy between fair and biased models. We present the more detailed results of target predictions in \textit{Appendix C}.

\begin{figure}[t]
\begin{center}
\includegraphics[width=\linewidth]{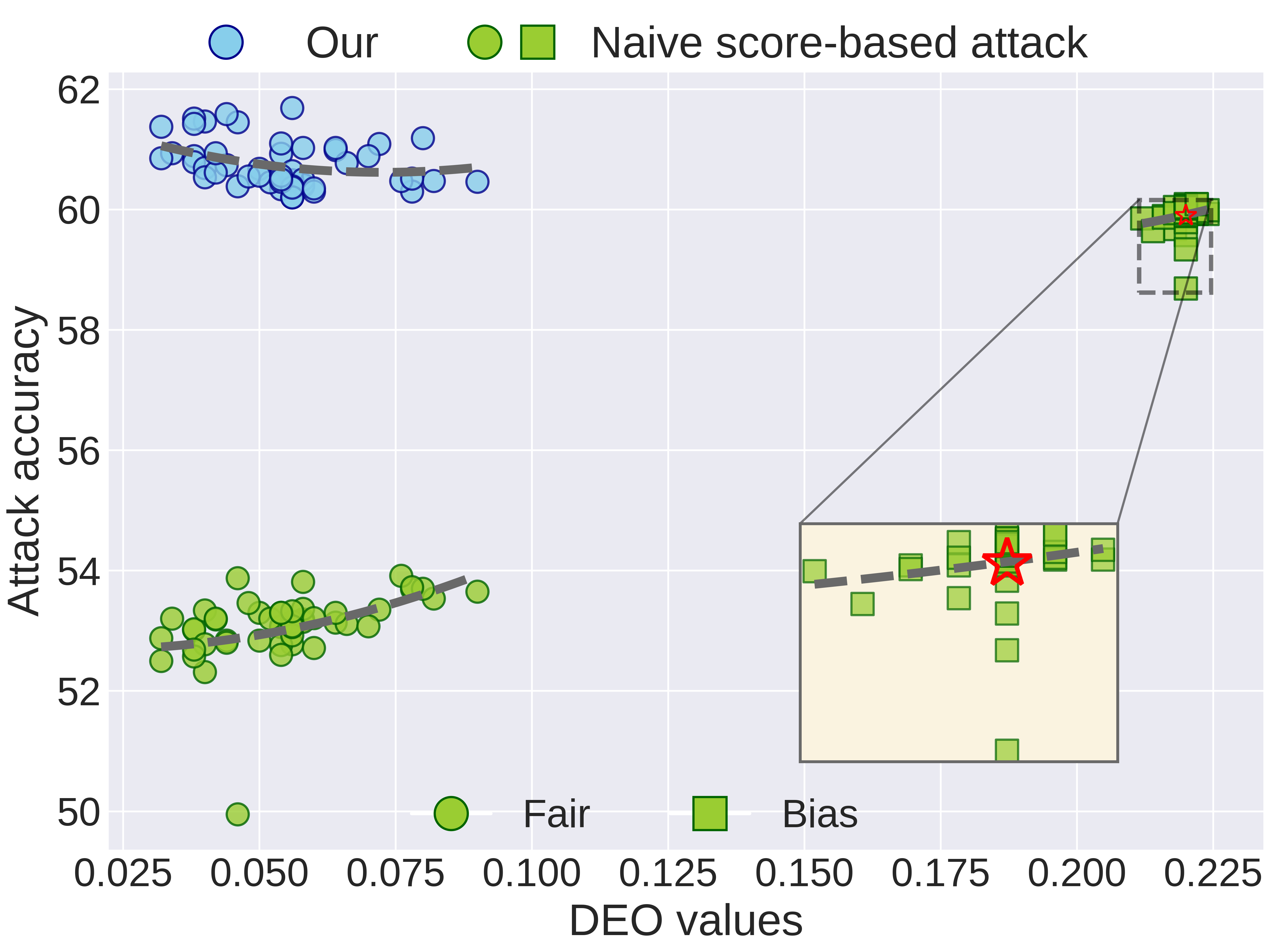}
\end{center}
\caption{Score-based attacks with models of varying fairness levels.}
\label{fig:compare}
\end{figure}

\noindent \textbf{Results with varying fairness levels.} Next, we attack models of different fairness performances. Specifically, we consider the case of CelebA (T=s/S=g) and conduct score-based attacks on biased and fair models of different DEO values. Figure~\ref{fig:compare} presents the results. For FD-MIA, we utilize prediction results from multiple fair models and one biased one, which is indicated by a red star in the figure. We further adopt dashed gray lines to outline the trend. 

The figure illustrates that attack accuracy decreases for both biased and fair models as the DEO value decreases. The results indicate that models with stronger fairness interventions exhibit more robustness against existing MIAs. While achieving improved fairness, these models lower their confidence scores, making the attacks more challenging. In contrast, FD-MIA, which exploits discrepancies in fairness, achieves superior attack performance. Particularly, larger fairness discrepancies between the fair and biased models contribute to larger prediction gaps, leading to more powerful attacks with FD-MIA.

\begin{figure}[t]
\begin{center}
\includegraphics[width=\linewidth]{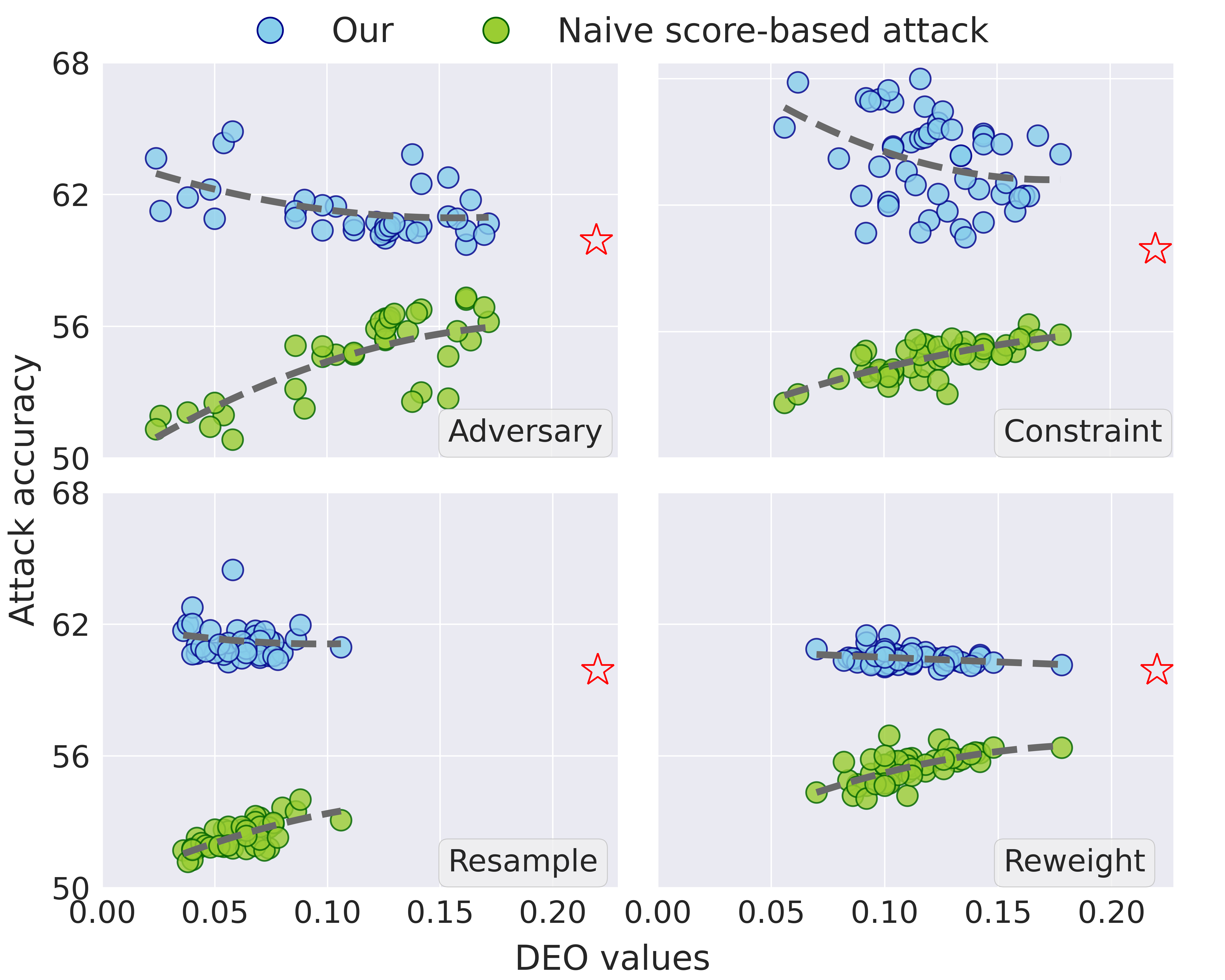}
\end{center}
\caption{Score-based attacks on models with different fairness intervention methods.}
\label{fig:fair_methods}
\end{figure}

\noindent \textbf{Results with different fairness approaches.} In this part, we evaluate our findings with various fairness approaches, including data sampling, reweighting, adversarial training, and constraint-based approaches. In the experiments, we adopt the implementations of these approaches from~\cite{wangMitigatingBiasFace,han2023ffb}. 
Similarly, we focus on the case of CelebA (T=s/S=g), and Figure~\ref{fig:fair_methods} presents the results. The figure shows reduced DEO values after fairness interventions, indicating the effectiveness of these approaches.

For attack results, the naive score-based attacks exhibit degraded performance with fair models for all fairness approaches. The attack accuracy drops as the DEO values reduce. The results align with our previous findings, where fairness interventions introduce some robustness to MIAs. Notably, the drops are more pronounced with the adversarial training and constraint approaches. We believe this is due to the more substantial trade-offs between fairness and utility inherent to the approaches. 

In contrast, for all approaches, FD-MIA achieved higher attack accuracy with fair models compared to biased ones. Similarly, the attack performance improves when the fairness discrepancy enlarges as FD-MIA explores the prediction gaps. The experiments demonstrate our findings and the proposed method with various representative fairness approaches. The results indicate fairness interventions can impose real privacy threats. We present more detailed results in \textit{Appendix D}.

We further conduct more diverse ablation studies, including attacks with different model structures (\textit{Appendix E}), attacks with non-binary classifiers (\textit{Appendix F}), attacks with tabular data (\textit{Appendix G}), and attacks with varying skewed data distributions (\textit{Appendix H}). We present detailed results in the Appendix, and all results demonstrate our findings and the proposed method. Thus, we conclude that FD-MIA is effective in attacking fair models and poses real threats to model privacy.

\section{Mitigation}
We further outline two potential defense mechanisms for the proposed attack method: 

\noindent \textbf{Restricting information access} involves limiting the adversary's access to the crucial information required for the attack methods. 
For instance, 
the defense can choose only to publish the predicted labels while withholding confidence scores, 
which are the essential information for membership inference attack (MIA) methods. 
Moreover, 
the defense may opt to release only the prediction results from fair models to reduce potential privacy leakage.

\noindent \textbf{Differential privacy} (DP), as introduced in~\cite{dwork2006calibrating}, imposes a constraint on the ability to distinguish between two neighboring datasets that differ by only a single data sample. It has been established as an effective countermeasure against MIAs. In our study, we employ the differentially private stochastic gradient descent (DP-SGD)~\cite{abadi2016deep} to defend against the attacks considering the models for the CelebA (T=s/S=g) in Table~\ref{table:dual_results}.
Table~\ref{table:dp_results} shows the attack results with the existing methods and the proposed FD-MIA. 
When DP noise is integrated, the attack accuracy is significantly reduced, demonstrating the effectiveness of the DP-SGD.
Notably, with the same magnitude of noise, our proposed ($\text{Our}_s$, $\text{Our}_l$) still outperforms existing attacks.
This implies that more noise is needed to attain comparable defense performance with the existing attacks. It also indicates the superior attack performance of the proposed FD-MIA compared to the existing ones. 
We further present the defense results for fairness performance, as well as different values of the DP budget $\epsilon$ in \textit{Appendix I}. 

\begin{table}[t]
\centering 
  \begin{tabular}{ c c c c}
  \toprule {Models}&
   {Acc} &
   {AUC} &
   {TPR@FPR} 
   \\
   \midrule
   {$\text{Fair}_{s}$} &50.8 ($\downarrow$ 2.4) &51.2 ($\downarrow$ 3.6) & 0.0 ($\downarrow$ 0.04) \\
   {$\text{Our}_{s}$} &53.4 ($\downarrow$ 7.2) &55.8 ($\downarrow$ 10.0) &0.0 ($\downarrow$ 0.3) \\
   \midrule
   {$\text{Fair}_{l}$} &50.5 ($\downarrow$ 0.3) &49.8 ($\downarrow$ 0.5) &0.1 ($\downarrow$ 0.1) \\
   {$\text{Our}_{l}$} &51.4 ($\downarrow$ 3.3) &51.2 ($\downarrow$ 6.1) &0.1 ($\downarrow$ 1.1) \\
   \bottomrule
  \end{tabular}
  \caption{DP-SGD results with $\delta=10^{-5}, \epsilon=0.85$ in (\%).}
  \label{table:dp_results}
\end{table}


\section{Discussions}
\subsection{Influential factors}
The proposed FD-MIA exploits the prediction gaps between biased and fair models to achieve enhanced MIA results. As indicated in Figures 8 and 9 of the main paper, the attack results can be influenced by two factors: the adopted fair model and the fairness method used.

In the main experiments, we considered fair models without severe accuracy degradation after fairness interventions. This led to smaller prediction gaps between the fair and biased models, resulting in relatively modest improvements in the attack performance.

Figure~\ref{fig:compare} shows that FD-MIA adopts fair models with varying fairness performance. We observe that fairer models generally lead to better attack results. This is due to the enlarged prediction gaps between the biased and fair models.

Figure~\ref{fig:fair_methods} considers different fairness methods, which result in different attack results. This discrepancy indicates that the fairness methods can also influence the attack results. Notably, the adversarial training and constraint approaches lead to more pronounced prediction gaps between biased and fair models, resulting in enhanced attack results. This may be due to the more substantial trade-offs between fairness and model utility inherent to these approaches.

\subsection{Ethical Considerations}
We reveal a concerning tension between the pursuit of algorithmic fairness and the preservation of individual privacy. While fairness interventions have shown promise in mitigating biased predictions, our results demonstrate that these techniques can inadvertently introduce new vulnerabilities to membership inference attacks, posing significant risks to model privacy.

This trade-off between fairness and privacy raises important ethical questions that warrant careful examination. On one hand, there is a moral imperative to ensure machine learning models do not perpetuate harmful biases and discriminatory practices
On the other hand, the privacy of individuals whose data is used to train these models is also an essential ethical consideration.

To explore the trade-offs between fairness and privacy, we measure model fairness using different metrics and assess the model's privacy performance with MIAs. These practices can be adopted to evaluate deployed models. The Model developers should engage with ethicists, policymakers, and affected communities to ensure that there are mechanisms in place to hold these systems accountable for their fairness and privacy implications.

To mitigate the privacy risks while maintaining fairness, we encourage the adoption of techniques like differential privacy (DP-SGD). By introducing carefully calibrated noise into the training process, differential privacy can help preserve the privacy of individuals whose data is used to train the models without significantly compromising their fairness performance.

Finally, the trade-off between fairness and privacy is a complex issue that requires nuanced ethical deliberation. By incorporating these key considerations, we aim to provide practitioners with a more holistic understanding of the challenges and guide them in making informed decisions that uphold both ethical principles to the greatest extent possible.

\section{Conclusions}
This paper evaluates the privacy risks of fairness interventions by employing membership inference attacks (MIAs). 
Surprisingly, 
our findings reveal that fair models exhibit unexpected robustness against existing MIAs.
However, 
these attacks are inefficient against binary classifiers as the trained attack models degrade into simple threshold models. 
We then proposed a novel attack method named FD-MIA, 
which leverages predictions from both biased and fair models to exploit the prediction gaps between member and non-member data. 
It can be integrated into existing attacks and pose substantial threads to model privacy with superior attack performance across different attack methods and metrics.
We conduct experiments across six datasets, three attack methods, and five representative fairness approaches involving a total of $32$ settings and more than $160$ different models. All results demonstrate our findings and the performance of the proposed attack. 
While our method primarily targets fair binary classifiers, 
the attack method can be extended to more generic models with fairness disparities. 
Our results shed new light on the privacy leakage risks associated with fairness methods, 
emphasizing the necessity for careful consideration in the design and deployment of trustworthy models. 

\bibliographystyle{named}
\bibliography{bib.bib,additional.bib}
\clearpage

\appendix
\section{Detailed experiment settings}
\begin{table*}[t]
\centering 
  \begin{tabular}{c c c c c c c c c}
  \toprule
  \multirow{2}{*}{Settings}     &  \multirow{2}{*}{Models}  & \multicolumn{3}{c}{Target learning results}                                & \multicolumn{3}{c}{Attack results}                      \\
                 \cmidrule{3-8}
                      &  & $\text{Acc}_\text{t}$ $\uparrow$ & $\text{BA}$ $\downarrow$ & $\text{DEO}$ $\downarrow$ & $\text{Acc}_\text{a}$ $\uparrow$ & $\text{AUC}_\text{a}$ $\uparrow$          & TPR@FPR  $\uparrow$      \\

    \midrule 
            \multirow{6}{*}{\parbox{1.5cm}{\centering CelebA \\(T=s/S=g)}}
            &{$\text{Bias}_{s}$} & 87.6 ($\pm$0.0) &7.7 ($\pm$0.1) & 21.7 ($\pm$0.0) &59.8 ($\pm$0.0) &62.8 ($\pm$0.6) & 0.0 ($\pm$0.0)   \\
            &{$\text{Fair}_{s}$} &90.5 ($\pm$0.0) &2.5 ($\pm$0.9) &5.6 ($\pm$1.0) &53.2 ($\pm$1.1) &54.8 ($\pm$0.4) & 0.0 ($\pm$0.0)  \\
            &{$\text{Our}_{s}$} &- &- &- &\textbf{60.6 ($\pm$0.2)} &\textbf{65.8 ($\pm$0.6)} & \textbf{0.3 ($\pm$0.1)}  \\
    \cmidrule{2-8}
            &{$\text{Bias}_{l}$} & 87.6 ($\pm$0.0) &7.7 ($\pm$0.1) & 21.7 ($\pm$0.0) &51.5 ($\pm$0.0) &51.4 ($\pm$0.2) & 0.6 ($\pm$0.1)  \\
            &{$\text{Fair}_{l}$} &90.5 ($\pm$0.0) &2.5 ($\pm$0.9) &5.6 ($\pm$1.0) &50.8 ($\pm$0.5) &50.3 ($\pm$0.3) & 0.2 ($\pm$0.0) \\
            &{$\text{Our}_{l}$} &- &- &- &\textbf{54.7 ($\pm$0.7)} &\textbf{57.3 ($\pm$1.0)} & \textbf{1.2 ($\pm$0.3)}  \\
    \midrule
            \multirow{6}{*}{\parbox{1.5cm}{\centering UTKFace \\(T=r/S=g)}}
            &{$\text{Bias}_{s}$} & 87.4 ($\pm$0.4) &3.6 ($\pm$0.3) &14.2 ($\pm$0.2)&58.5 ($\pm$0.1) &58.9 ($\pm$0.4) &0.0 ($\pm$0.0) \\
            &{$\text{Fair}_{s}$} &89 ($\pm$0.5) &0.8 ($\pm$0.1) &6.3($\pm$0.2)&52.6 ($\pm$0.1) &52.8 ($\pm$0.7) &0.1 ($\pm$0.1) \\
            &{$\text{Our}_{s}$} &-&-&- &\textbf{60.2 ($\pm$2.5)} &\textbf{62.1 ($\pm$1.9)} &\textbf{0.2 ($\pm$0.1)} \\
    \cmidrule{2-8}
            &{$\text{Bias}_{l}$} & 87.4 ($\pm$0.4) &3.6 ($\pm$0.3) &14.2 ($\pm$0.2)&55.4 ($\pm$0.3) &51.5 ($\pm$0.3) &0.9 ($\pm$0.4) \\
            &{$\text{Fair}_{l}$} &89 ($\pm$0.5) &0.8 ($\pm$0.1) &6.3($\pm$0.2)&53.2 ($\pm$0.2) &47.6 ($\pm$0.4) &0.7 ($\pm$0.3) \\
            &{$\text{Our}_{l}$} &-&-&-&\textbf{55.9 ($\pm$0.3)} &\textbf{52.2 ($\pm$0.2)} &\textbf{1.7 ($\pm$0.3)} \\
    \midrule
            \multirow{6}{*}{\parbox{1.5cm}{\centering FaceFace \\(T=r/S=g)}}
            &{$\text{Bias}_{s}$} & 87.2 ($\pm$0.0) &7.7 ($\pm$0.0) &22.2 ($\pm$0.4)&63.6 ($\pm$0.2) &66.4 ($\pm$0.0) &0.0 ($\pm$0.0) \\ 
            &{$\text{Fair}_{s}$} &87.6 ($\pm$0.1) &1.9 ($\pm$0.6) &3.9($\pm$0.2)&63.3 ($\pm$0.3) &66.2 ($\pm$0.2) &0.0 ($\pm$0.0) \\
            &{$\text{Our}_{s}$} &-&-&-&\textbf{65.2 ($\pm$0.1)} &\textbf{66.8 ($\pm$0.7)} &\textbf{0.2 ($\pm$0.0)} \\
    \cmidrule{2-8}
            &{$\text{Bias}_{l}$} & 87.2 ($\pm$0.0) &7.7 ($\pm$0.0) &22.2 ($\pm$0.4)&60.2 ($\pm$0.7) &61.7 ($\pm$1.2) &1.3 ($\pm$0.5) \\  
            &{$\text{Fair}_{l}$} &87.6 ($\pm$0.1) &1.9 ($\pm$0.6) &3.9($\pm$0.2)&56.7 ($\pm$0.2) &57.2 ($\pm$0.3) &0.9 ($\pm$0.1) \\
            &{$\text{Our}_{l}$} &-&-&-&\textbf{62.3 ($\pm$0.2)} &\textbf{63.2 ($\pm$0.4)} &\textbf{2.3 ($\pm$0.3)} \\   
    \bottomrule
  \end{tabular}
  \caption{Target learning and attack results for the \textit{gender} attribute in (\%).}
  \label{table:result_gender}
\end{table*}

\begin{table*}[ht]
\centering 
  \begin{tabular}{c l l l l l l l}
  \toprule 
                 \multirow{2}{*}{MIAs}     &  \multirow{2}{*}{\parbox{1.5cm}{\centering Fairness \\approaches}}  & \multicolumn{3}{c}{Target learning for CelebA (T=s/S=g)}                                & \multicolumn{3}{c}{Attack results}                      \\
                 \cmidrule{3-8}
                      &  & $\text{Acc}_\text{t}$ $\uparrow$ & $\text{BA}$ $\downarrow$ & $\text{DEO}$ $\downarrow$ & $\text{Acc}_\text{a}$ $\uparrow$ & $\text{AUC}_\text{a}$ $\uparrow$          & TPR@FPR  $\uparrow$      \\
    \midrule
    \multirow{4}{*}{\parbox{1.5cm}{\centering Score \\based}}
            &{Adversarial} & 85.2 ($\pm$0.0) &3.4 ($\pm$1.2) &11.3 ($\pm$3.1) &54.9 ($\pm$1.7) &57.3 ($\pm$0.0) &0.1 ($\pm$0.0)  \\
            &{Constraint} &87.8 ($\pm$2.1) &6.2 ($\pm$2.3) &12.3($\pm$2.6)&54.7 ($\pm$0.8) &57.1 ($\pm$0.0) &0.1 ($\pm$0.0) \\
            &{Resample} &89.9 ($\pm$0.0) &2.6 ($\pm$0.7) &6.1($\pm$0.5)&52.3 ($\pm$0.7) &53.6 ($\pm$0.0) &0.1 ($\pm$0.0) \\
            &{Reweight} &89.5 ($\pm$1.0) &2.7 ($\pm$0.9) &6.1($\pm$1.5)&55.4 ($\pm$0.6) &57.6 ($\pm$0.0) &0.1 ($\pm$0.0) \\
    \midrule
    \multirow{4}{*}{Our}
            &{Adversarial} & 85.2 ($\pm$0.00) &3.4 ($\pm$1.2) &11.3 ($\pm$3.1)&61.9 ($\pm$2.7) &63.7 ($\pm$2.2) &3.3 ($\pm$0.6) \\
            &{Constraint} &87.8 ($\pm$0.02) &6.2 ($\pm$2.3) &12.3($\pm$2.6)&64.3 ($\pm$2.3) &66.2 ($\pm$2.1) &3.9 ($\pm$0.7) \\
            &{Resample} &89.9 ($\pm$0.00) &2.6 ($\pm$0.7) &6.1($\pm$0.5)&61.2 ($\pm$0.0) &65.6 ($\pm$0.6) &0.2 ($\pm$0.1) \\
            &{Reweight} &89.5 ($\pm$0.0) &2.7 ($\pm$0.9) &6.1($\pm$1.5)&60.9 ($\pm$0.2) &63.3 ($\pm$02) &0.2 ($\pm$0.1) \\
    \bottomrule
  \end{tabular}
  \caption{Detailed results with different fairness approaches in (\%).}
  \label{table:fair_appoaches}
\end{table*}

\begin{table*}[t]
\centering 
  \begin{tabular}{c c l l l l l l l}
  \toprule
\multirow{2}{*}{Settings}     &  \multirow{2}{*}{Models}  & \multicolumn{3}{c}{Target learning results}                                & \multicolumn{3}{c}{Attack results}                      \\
                 \cmidrule{3-8}
                      &  & $\text{Acc}_\text{t}$ $\uparrow$ & $\text{BA}$ $\downarrow$ & $\text{DEO}$ $\downarrow$ & $\text{Acc}_\text{a}$ $\uparrow$ & $\text{AUC}_\text{a}$ $\uparrow$          & TPR@FPR  $\uparrow$      \\
    \midrule 
            \multirow{6}{*}{\parbox{1.5cm}{\centering CelebA \\(T=s/S=h)}}
            &{$\text{Bias}_{s}$} & 89.9 ($\pm$0.1) &2.5 ($\pm$0.0) &10.4 ($\pm$0.1) &55.1 ($\pm$0.1) &56.3 ($\pm$0.0) &0.1 ($\pm$0.1) \\
            &{$\text{Fair}_{s}$} & 90.1 ($\pm$0.4) &0.9 ($\pm$0.2) &3.7($\pm$0.5) &52.6 ($\pm$0.1) &52.7 ($\pm$0.0) &0.0 ($\pm$0.0) \\
            &{$\text{Our}_{s}$} &- &- &- &\textbf{56.9 ($\pm$0.3)} &\textbf{59.6 ($\pm$0.0)} &\textbf{0.2 ($\pm$0.0)} \\
    \cmidrule{2-8}
            &{$\text{Bias}_{l}$} & 89.9 ($\pm$0.1) &2.5 ($\pm$0.0) &10.4 ($\pm$0.1) &52.1 ($\pm$0.3) &52.0 ($\pm$0.8) &0.3 ($\pm$0.1) \\
            &{$\text{Fair}_{l}$} & 90.1 ($\pm$0.4) &0.9 ($\pm$0.2) &3.7($\pm$0.5)&51.0 ($\pm$0.3) &50.5 ($\pm$0.6) &0.1 ($\pm$0.1) \\
            &{$\text{Our}_{l}$} &- &- &- &\textbf{55.4 ($\pm$0.5)} &\textbf{57.7 ($\pm$0.1)} &\textbf{0.8 ($\pm$0.1)} \\
    \midrule
            \multirow{6}{*}{\parbox{1.5cm}{\centering CelebA \\(T=s/S=m)}}
            &{$\text{Bias}_{s}$} & 88.6 ($\pm$0.1) &3.5 ($\pm$0.0) &14.6 ($\pm$0.1) &57.4 ($\pm$0.1) &58.1 ($\pm$0.0) &0.0 ($\pm$0.0) \\
            &{$\text{Fair}_{s}$} & 90.5 ($\pm$0.3) &0.8 ($\pm$0.1) &2.4($\pm$0.6) &53.1 ($\pm$0.3) &52.0 ($\pm$0.0) &0.0 ($\pm$0.0) \\
            &{$\text{Our}_{s}$} &-&-&- &\textbf{59.6 ($\pm$0.2)} &\textbf{63.2 ($\pm$0.4)} &\textbf{0.2 ($\pm$0.0)} \\
    \cmidrule{2-8}
            &{$\text{Bias}_{l}$} & 88.6 ($\pm$0.1) &3.5 ($\pm$0.0) &14.6 ($\pm$0.1)&51.6 ($\pm$0.4) &51.4 ($\pm$0.6) &0.4 ($\pm$0.1) \\
            &{$\text{Fair}_{l}$} & 90.5 ($\pm$0.3) &0.8 ($\pm$0.1) &2.4($\pm$0.6)&50.7 ($\pm$0.5) &49.9 ($\pm$0.9) &0.1 ($\pm$0.1) \\
            &{$\text{Our}_{l}$} &-&-&-&\textbf{54.2 ($\pm$0.5)} &\textbf{55.7 ($\pm$1.1)} &\textbf{0.6 ($\pm$0.1)} \\
    \midrule
            \multirow{6}{*}{\parbox{1.5cm}{\centering UTKFace \\(T=g/S=r)}}
            &{$\text{Bias}_{s}$} & 80.8 ($\pm$0.1) &8.8 ($\pm$0.6) &31.9 ($\pm$1.4)&64.0 ($\pm$1.3) &66.9 ($\pm$0.4) &0.0 ($\pm$0.0) \\ 
            &{$\text{Fair}_{s}$} &86.3 ($\pm$0.4) &2.8 ($\pm$0.4) &14.3($\pm$0.4)&55.3 ($\pm$0.8) &57.2 ($\pm$1.0) &0.0 ($\pm$0.0) \\
            &{$\text{Our}_{s}$} &-&-&-&\textbf{66.7 ($\pm$0.1)} &\textbf{67.8 ($\pm$0.7)} &\textbf{0.3 ($\pm$0.1)} \\
    \cmidrule{2-8}
            &{$\text{Bias}_{l}$} & 80.8 ($\pm$0.1) &8.8 ($\pm$0.6) &31.9 ($\pm$1.4) &55.5 ($\pm$0.2) &52.4 ($\pm$0.4) &1.4 ($\pm$0.1) \\ 
            &{$\text{Fair}_{l}$} &86.3 ($\pm$0.4) &2.8 ($\pm$0.4) &14.3($\pm$0.4)&53.8 ($\pm$0.1) &49.7 ($\pm$0.3) &0.9 ($\pm$0.1) \\
            &{$\text{Our}_{l}$} &-&-&-&\textbf{56.2 ($\pm$0.2)} &\textbf{53.6 ($\pm$0.4)} &\textbf{2.1 ($\pm$0.3)} \\ 
    \midrule
            \multirow{6}{*}{\parbox{1.5cm}{\centering FaceFace \\(T=g/S=r)}}
            &{$\text{Bias}_{s}$} & 90.5 ($\pm$0.3) &12.5 ($\pm$0.7) &5.3 ($\pm$1.1) &75.5 ($\pm$1.7) &76.7 ($\pm$0.4) &0.0 ($\pm$0.0) \\
            &{$\text{Fair}_{s}$} &92.0 ($\pm$0.3) &5.1 ($\pm$2.0) &4.5($\pm$1.1) &73.2 ($\pm$0.9) &75.5 ($\pm$0.3) &0.0 ($\pm$0.0) \\
            &{$\text{Our}_{s}$} &-&-&-&\textbf{77.0 ($\pm$0.3)} &\textbf{78.4 ($\pm$0.0)} &\textbf{0.7 ($\pm$0.1)} \\
    \cmidrule{2-8}
            &{$\text{Bias}_{l}$} & 90.5 ($\pm$0.3) &12.5 ($\pm$0.7) &5.3 ($\pm$1.1)&73.2 ($\pm$0.3) &74.2 ($\pm$0.4) &1.5 ($\pm$0.1) \\
            &{$\text{Fair}_{l}$} &92.0 ($\pm$0.3) &5.1 ($\pm$2.0) &4.5($\pm$1.1)&70.4 ($\pm$0.1) &72.1 ($\pm$0.1) &0.6 ($\pm$0.4) \\
            &{$\text{Our}_{l}$} &-&-&-&\textbf{75.2 ($\pm$1.1)} &\textbf{76.4 ($\pm$1.3)} &\textbf{2.9 ($\pm$0.7)} \\
    \bottomrule
  \end{tabular}
  \caption{Target learning and attack results for different attributes in (\%).}
  \label{table:attributes}
\end{table*}

\begin{table*}[t]
\centering 
  \begin{tabular}{c c l l l l l l l}
  \toprule

   \multirow{2}{*}{Structures}     &  \multirow{2}{*}{Models}  & \multicolumn{3}{c}{Target learning results}                                & \multicolumn{3}{c}{Attack results}                      \\
                 \cmidrule{3-8}
                      &  & $\text{Acc}_\text{t}$ $\uparrow$ & $\text{BA}$ $\downarrow$ & $\text{DEO}$ $\downarrow$ & $\text{Acc}_\text{a}$ $\uparrow$ & $\text{AUC}_\text{a}$ $\uparrow$          & TPR@FPR  $\uparrow$      \\ 
    \midrule 
            \multirow{3}{*}{\centering ResNet18}
            &{$\text{Bias}$} & 78.5 ($\pm$0.1) &12.8 ($\pm$0.1) &35.5 ($\pm$0.2)&59.6 ($\pm$0.6) &69.2 ($\pm$0.0) &0.0($\pm$0.0) \\
            &{$\text{Fair}$} &90.0 ($\pm$0.3) &2.7 ($\pm$0.9) &6.5($\pm$0.7)&54.2 ($\pm$0.1) &56.5 ($\pm$0.0) &0.0 ($\pm$0.0) \\
            &{$\text{Our}$} &- &- &- &\textbf{64.5 ($\pm$0.0)} &\textbf{73.5 ($\pm$0.0)} &\textbf{0.3 ($\pm$0.0)} \\
    \midrule
            \multirow{3}{*}{\centering VGG}
            &{$\text{Bias}$} & 87.4 ($\pm$0.4) &7.2 ($\pm$0.2) &20.7 ($\pm$0.4)&55.2 ($\pm$0.2) &61.9 ($\pm$0.0) &0.0 ($\pm$0.0) \\
            &{$\text{Fair}$} & 93.4 ($\pm$0.3) &2.3 ($\pm$0.5) &4.9($\pm$0.3)&52.2 ($\pm$0.8) &54.6 ($\pm$0.01) &0.0 ($\pm$0.0) \\
            &{$\text{Our}$} &-&-&- &\textbf{59.6 ($\pm$0.2)} &\textbf{63.2 ($\pm$0.4)} &\textbf{0.2 ($\pm$0.0)} \\
    \bottomrule
  \end{tabular}
  \caption{Attack results with different model structures in (\%).}
  \label{table:model}
\end{table*}

\subsection{Datasets}
For the CelebA~\cite{CelebAMask-HQ} data, we evaluate our findings with various considered sensitive attributes and focus on smiling predictions. The UTKFace dataset~\cite{geraldsutkface} consists of more than $20,000$ facial images, which are annotated with ethnicities, age, and gender information. The dataset includes five main ethnicity groups, namely \textit{White, Black, Asian, Indian}, and \textit{Others} (including \textit{Hispanic}, \textit{Latino}). As the group \textit{White} dominates the dataset, we group the remaining races into the \textit{Others} category. In the experiments, we consider the binary attribute values for the attribute of \textit{race}. The FairFace dataset~\cite{karkkainen2021fairface} comprises over $100,000$ facial images annotated with information on races, genders, and ages. Similar to the UTKFace dataset, we group the data into subgroups and consider the binary attribute values for \textit{race}.

\subsection{Training data} For our experiments, membership inference attack (MIA) evaluation requires splitting datasets into training and non-training subsets. To induce biased predictions, we randomly sample the training data to skew the distribution across specified sensitive attributes. Specifically, we set a highly imbalanced ratio of $9:1$ between the majority and minority groups for the sensitive attribute (e.g., $90\%$ male data and $10\%$ female data). Meanwhile, we maintain a balanced $50-50\%$ data distribution for the target learning (e.g., $50\%$ smiling and $50\%$ non-smiling). We also maintain a balanced data distribution for member and non-member data for training attack models. We further present ablation studies with varying skewed data distributions to evaluate our findings and the proposed method in \textit{Appendix H}. 

\subsection{Models} 
In the experiments, we utilize a 6-layer deep model as the target model, comprising three consecutive CNN layers followed by three linear layers. To enforce fairness in the predictions, we introduce an additional attribute prediction head on top of the target prediction head, following the detailed approach described in \cite{Wang2020TowardsFI}. For mixup operations, we interpolate the input data from different subgroups following the setting in~\cite{mroueh2021fair,du2021fairness}. We further evaluate our methods with different prediction models and various fairness approaches in \textit{Appendix E} and \textit{Appendix D}.

\subsection{Fairness metrics}
Given biased models, 
we consider a sensitive attribute $s \in S$ and subgroups $\{s_0, s_1\}$ with binary attribute values $\{0,1\}$. 
As the prediction target $Y$ and sensitive attributes are irrelevant, 
the model prediction $ \widetilde{Y}$ and $S$ should be independent. 
To measure the unfairness, 
we adopt bias amplification~\cite{zhao-etal-2017-men} and equalized odds~\cite{Hardt:2016wv} as the fairness metrics.

\textbf{Bias amplification} (BA)~\cite{zhao-etal-2017-men} requires equal results of true positive predictions across all subgroups and can be written as
\begin{equation}
        \frac{1}{2} \frac{\left|\mathrm{TP}_{s_0}- \mathrm{TP}_{s_1}\right|}{\mathrm{TP}_{s_0}+\mathrm{TP}_{s_1}},
\end{equation}
where $\text{TP}$ presents the true positive value of predictions.

\textbf{Equalized odds} (EO)~\cite{Hardt:2016wv} requires equal results of true positive rates (TPRs) and false positive rates (FPRs) for different groups. The fairness metric can be defined as:
\begin{equation}
        P{\{\widetilde{Y}|Y, S\}}; Y=\{0,1\} , S=\{s_0, s_1\}.
\end{equation}
We report the discrimination level with the difference of BA and EO across subgroups for fairness evaluations.

\section{Attacks with the \textit{gender} attribute}
Table~\ref{table:result_gender} presents the detailed results for attacks considering the sensitive attribute of \textit{gender} across different datasets. Specifically, 
\textbf{CelebA with (T=s/S=g)} presents the experiments with the CelebA dataset considering T=s/S=g; \textbf{UTKFace with (T=r/S=g)} is for the UTKFace dataset considering T=r/S=g; \textbf{FaceFace with (T=r/S=g)} indicates experiments for the FaceFace dataset considering T=r/S=g. We first present the results for learning targets ($\text{Acc}_\text{t}$) and then present the results of fairness metrics ($\text{BA}$, $\text{DEO}$). We also present the attack results with accuracy and AUC values ($\text{Acc}_\text{a}$, $\text{AUC}_\text{a}$, and TPR@FPR). For attack results, we integrate the proposed method with the score-based attack method ($s$)~\cite{shokri2017membership} and the reference-based attack method of LiRA ($l$)~\cite{carlini2022membership}. All experimental results demonstrate our findings and the effectiveness of the proposed method.

\section{Attacks with different attributes}
We consider different learning targets and sensitive attributes. Specifically, for \textbf{CelebA with (T=s/S=h)}, we consider \textit{wavy hair} as the sensitive attribute and smiling as prediction targets; for \textbf{CelebA with (T=s/S=m)}, we consider \textit{heavy make} as the sensitive attribute and smiling as prediction targets; for \textbf{UTKFace with (T=g/S=r)}, we consider \textit{race} as the sensitive attribute and gender as prediction targets; for \textbf{FaceFace with (T=g/S=r)}, we consider \textit{race} as the sensitive attribute and gender as prediction targets. Table~\ref{table:attributes} presents the results for the target learning and the attack results.

\section{Attacks with different fairness approaches}
In this part, we attack fair models with the score-based attack method and the proposed method considering the biased data in CelebA (T=s/S=g). Differently, to obtain the fair models, we apply the following debiasing approaches, including Adversarial training, Fair Constraint, Re-sampling, and Re-weighting methods. Table~\ref{table:fair_appoaches} presents the results for the target learning and the results for the attacks.

\section{Attacks with different model structures}
We evaluate the performance of the proposed methods considering different model structures of ResNet18~\cite{He2016DeepRL} and VGG~\cite{Simonyan2015VeryDC}. Table~\ref{table:model} presents the results of target predictions as well as attack results. In the experiments, we adopt the CelebA dataset and consider \textit{gender} as sensitive attributes to predict the attribute of smiling. The results indicate that the proposed method can achieve enhanced attack results with different model structures. In contrast, existing attack methods consistently suffer inefficient attacks with inferior performance.

\section{Attacks with non-binary classifiers}
Previous experiments demonstrate the proposed FD-MIA in binary classifications. In this part, we extend the proposed method to multi-class predictions. Specifically, we consider the dataset of CIFAR-10S from ~\cite{Wang2020TowardsFI}, which includes CIFAR-10~\cite{krizhevsky2009learning} images in both colored and grayscale formats. The setting allows for assessing model fairness performance considering the sensitive attribute of \textit{color}. In the experiments, following the setting in~\cite{wangMitigatingBiasFace}, we set the training data from the classes of \textit{automobile, cat, dog, horse, truck} to be dominated by images with color and the remainder by the grayscale ones. For each class, the training data are split into the subgroups of color ($80$\% of the entire data) and gray ($20$\% of the entire data). 

\begin{table*}[ht]
\centering 
\begin{tabular}{ccccccc}
\toprule 
\multirow{2}{*}{Model} & \multicolumn{3}{c}{Target learning for CIFAR-10S (S=color)} & \multicolumn{3}{c}{Attack results} \\
\cmidrule{2-7}
                       & $\text{Acc}_\text{t}$ $\uparrow$  & BA $\downarrow$ & DEO $\downarrow$  & $\text{Acc}_\text{a}$ $\uparrow$  & $\text{AUC}_\text{a}$ $\uparrow$   & TPR@FPR $\uparrow$ \\
\midrule
$\text{Bias}_{s}$                   & 80.5 ($\pm$0.0)   & 2.7 ($\pm$0.0) & 6.3 ($\pm$0.1)  & 65.2 ($\pm$0.3)    & 80.1 ($\pm$0.4)  & 0.7 ($\pm$0.1) \\
$\text{Fair}_{s}$                   & 77.7 ($\pm$0.2)  & 1.1 ($\pm$0.3)  & 2.0 ($\pm$0.2)  & 56.9 ($\pm$0.0)    & 59.1 ($\pm$0.1)  & 0.1 ($\pm$0.0) \\
$\text{Our}_{s}$                    & -   &-    & -    & \textbf{67.0 ($\pm$0.1)}    & \textbf{84.0 ($\pm$0.0)}  & \textbf{2.2 ($\pm$0.3)} \\
\midrule
$\text{Bias}_{l}$                   & 80.5 ($\pm$0.0) & 2.7 ($\pm$0.0)   & 6.3 ($\pm$0.1)  & 54.6 ($\pm$0.5)    & 54.3 ($\pm$1.1)  & 0.7 ($\pm$0.2) \\
$\text{Fair}_{l}$                   & 77.7 ($\pm$0.2)   & 1.1 ($\pm$0.3)  & 2.0 ($\pm$0.2)  & 50.5 ($\pm$0.5)    & 50.1 ($\pm$0.7)  & 0.1 ($\pm$0.0) \\
$\text{Our}_{l}$                    & -   &-    & -    & \textbf{65.2 ($\pm$0.7)}    & \textbf{72.7 ($\pm$0.8)}  & \textbf{2.4 ($\pm$0.4)} \\
\bottomrule
\end{tabular}
\caption{Attack results for non-binary classifications with CIFAR-10S in (\%).}
\label{table:cifar}
\end{table*}

\begin{table*}[ht]
\centering
\small
\begin{tabular}{lllllllllllllllll}
\toprule 
\multirow{2}{*}{Model} & \multicolumn{4}{c}{Adult (T=i/S=g)} & \multicolumn{4}{c}{Adult (T=i/S=g)} & \multicolumn{4}{c}{COMPAS (T=r/S=g)} & \multicolumn{4}{c}{COMPAS (T=r/S=s)} \\
\cmidrule{2-17}
                       & $\text{Acc}_{t}$    & DEO    & $\text{Acc}_{a}$   & TPR  & $\text{Acc}_{t}$    & DEO    & $\text{Acc}_{a}$   & TPR  & $\text{Acc}_{t}$    & DEO   & $\text{Acc}_{a}$    & TPR   & $\text{Acc}_{t}$    & DEO    & $\text{Acc}_{a}$    & TPR  \\
\midrule
Bias                   & 68.9   & 19.9   & 54.1  & 0.0  & 65.9   & 58.1   & 52.4  & 0.1  & 70.5   & 5.9   & 54.1   & 0.0   & 61.9   & 15.2   & 56.9   & 0.0  \\
Fair                   & 75.2   & 4.9    & 53.1  & 0.0  & 75.0   & 16.5   & 51.8  & 0.1  & 69.1   & 1.9   & 53.6   & 0.0   & 63.9   & 8.5    & 54.7   & 0.0  \\
Our                    & -      & -      & \textbf{55.0}  & \textbf{0.1}  & -      & -      & \textbf{53.7}  & \textbf{0.2}  & -      & -     & \textbf{54.5}   & 0.0   & -      & -      & \textbf{58.7}   & \textbf{0.2} \\
\bottomrule
\end{tabular}
\caption{Attack results for tabular data of Adult and COMPAS in (\%).}
\label{table:tabular}
\end{table*}

Table~\ref{table:cifar} presents the results for biased and fair models. As shown in the table, the fairness-enhanced model achieved smaller DEO values with slightly lower accuracy results compared to the biased ones. On the other hand, the biased models tend to have better attack results than the fair ones. However, our attack methods combined the prediction information of both models and achieved the best results in both score-based and LiRA-based attacks. The results indicate the effectiveness of the proposed method in multi-class predictions. 

\begin{table*}[t]
\centering
\begin{tabular}{@{}ccllllll@{}}
\toprule 
                 \multirow{2}{*}{Distributions}     &  Target   & \multicolumn{3}{c}{Target learning for CelebA (T=s/S=g)}                                & \multicolumn{3}{c}{Score-based attack results}                      \\
                 \cmidrule{3-8}
                      & Models     & $\text{Acc}_\text{t}$ $\uparrow$ & $\text{BA}$ $\downarrow$ & $\text{DEO}$ $\downarrow$ & $\text{Acc}_\text{a}$ $\uparrow$ & $\text{AUC}_\text{a}$ $\uparrow$          & TPR@FPR  $\uparrow$      \\
\midrule
\multirow{3}{*}{0.95} & Bias & 82.7 ($\pm$0.0)                  & 9.6 ($\pm$0.0)           & 31.0 ($\pm$0.0)           & 63.8 ($\pm$0.4)                  & 67.7 ($\pm$0.1) & 0.0 ($\pm$0.0) \\
                      & Fair & 89.3 ($\pm$0.3)                  & 3.3 ($\pm$0.8)           & 9.2 ($\pm$1.1)            & 54.9 ($\pm$ 0.4)                 & 58.2 ($\pm$0.3) & 0.0 ($\pm$0.0) \\
                      & Our  & -                                & -                        & -                         & \textbf{64.5 ($\pm$0.3)}                  & \textbf{69.4 ($\pm$0.2)} & \textbf{0.2 ($\pm$0.0)} \\
\midrule
\multirow{3}{*}{0.9}  & Bias & 87.6 ($\pm$0.0)                  & 7.7 ($\pm$0.1)           & 21.7 ($\pm$0.0)           & 59.8 ($\pm$0.0)                  & 62.8 ($\pm$0.6) & 0.0 ($\pm$0.0) \\
                      & Fair & 90.5 ($\pm$0.0)                  & 2.5 ($\pm$0.9)           & 5.6 ($\pm$1.0)            & 53.2 ($\pm$1.1)                  & 54.8 ($\pm$0.4) & 0.0 ($\pm$0.0) \\
                      & Our  & -                                & -                        & -                         & \textbf{60.6 ($\pm$0.2)}                  & \textbf{65.8 ($\pm$0.6)} & \textbf{0.3 ($\pm$0.1)} \\
\midrule
\multirow{3}{*}{0.85} & Bias & 88.6 ($\pm$0.3)                  & 4.8 ($\pm$0.2)           & 17.0 ($\pm$0.7)           & 59.6 ($\pm$0.7)                  & 60.7 ($\pm$0.0) & 0.0 ($\pm$0.0) \\
                      & Fair & 90.1 ($\pm$0.5)                  & 1.7 ($\pm$0.8)           & 3.8 ($\pm$0.8)            & 54.1 ($\pm$0.3)                  & 55.3 ($\pm$0.7) & 0.0 ($\pm$0.0) \\
                      & Our  & -                                & -                        & -                         & \textbf{60.3 ($\pm$0.1)}                  & \textbf{63.6 ($\pm$0.7)} & \textbf{0.3 ($\pm$0.1)} \\
\midrule
\multirow{3}{*}{0.8}  & Bias & 88.1 ($\pm$0.3)                  & 5.0 ($\pm$0.3)           & 12.4 ($\pm$0.6)           & 58.7 ($\pm$0.8)                  & 57.2 ($\pm$0.5) & 0.0 ($\pm$0.0) \\
                      & Fair & 90.5 ($\pm$0.3)                  & 1.9 ($\pm$0.6)           & 4.1 ($\pm$1.1)            & 54.2 ($\pm$0.2)                  & 54.3 ($\pm$0.5) & 0.0 ($\pm$0.0) \\
                      & Our  & -                                & -                        & -                         & \textbf{59.5 ($\pm$0.2)}                  & \textbf{61.9 ($\pm$0.1)} & \textbf{0.2 ($\pm$0.0)} \\
\midrule
\multirow{3}{*}{0.75} & Bias & 89.1 ($\pm$0.1)                  & 1.6 ($\pm$0.4)           & 10.1 ($\pm$0.7)           & 56.7 ($\pm$0.4)                  & 56.5 ($\pm$0.1) & 0.0 ($\pm$0.0) \\
                      & Fair & 90.1 ($\pm$0.4)                  & 0.8 ($\pm$0.3)           & 2.8 ($\pm$1.2)            & 53.8 ($\pm$0.3)                  & 53.9 ($\pm$0.4) & 0.0 ($\pm$0.0) \\
                      & Our  & -                                & -                        & -                         & \textbf{58.3 ($\pm$0.7)}                  & \textbf{61.1 ($\pm$0.9)} & \textbf{0.3 ($\pm$0.1)} \\
\midrule
\multirow{3}{*}{0.7}  & Bias & 92.0 ($\pm$0.3)                  & 1.9 ($\pm$0.0)           & 4.8 ($\pm$0.1)            & 56.9 ($\pm$1.4)                  & 55.7 ($\pm$1.5) & 0.0 ($\pm$0.0) \\
                      & Fair & 91.4 ($\pm$0.4)                  & 0.9 ($\pm$0.5)           & 2.7 ($\pm$0.6)            & 53.9 ($\pm$0.3)                  & 53.7 ($\pm$0.5) & 0.0 ($\pm$0.0) \\
                      & Our  & -                                & -                        & -                         & \textbf{57.3 ($\pm$0.3)}                  & \textbf{59.4 ($\pm$1.3)} & \textbf{0.3 ($\pm$0.0)}\\
\bottomrule
\end{tabular}
\caption{Attack results for different skewed distributions.}
\label{table_dist}
\end{table*}

\section{Attacks with tabular data}
The previous study~\cite{chang2021privacy} evaluated the privacy impact of fairness methods with tabular data. They focused on decision tree models and attacked fair models with score-based attack methods. Following the setting, we further evaluate our findings with the Adult dataset~\cite{asuncion2007uci} and the COMPAS dataset~\cite{larson2016compas}. While the Adult dataset is adopted for income predictions (high or low) with collected information such as age or race, the COMPAS dataset is for assessing recidivism. In the experiments, we consider the sensitive attributes of \textit{gender} and \textit{age} for the Adult dataset, while the attributes of \textit{gender} and \textit{race} for the COMPAS dataset. We present the results in Table~\ref{table:tabular}, where S presents the considered attributes. The results indicate fair models tend to have inferior attack results compared with biased ones, while the proposed attack method achieved the best results. The results demonstrate our findings and the proposed methods.

We notice that fair models tend to have better attack results than the biased ones in~\cite{chang2021privacy}, which differs from our observations. We believe this can be attributed to the different classification models. In their work, the target models are trained with decision trees and re-weighing methods~\cite{pmlr-v80-agarwal18a} are adopted to obtain fair models. Their results illustrate that the fairness method leads to decreased losses and increased confidence scores for both majority and minority subgroups, contributing to more successful attacks. In contrast, the deep models we adopt can achieve high confidence scores even without fairness interventions. The fairness approaches decrease the scores for the majority group to pursue fairer predictions, leading to less effective attacks in our case. 

\begin{figure}[ht]
    \centering
    \includegraphics[width=.5\textwidth]{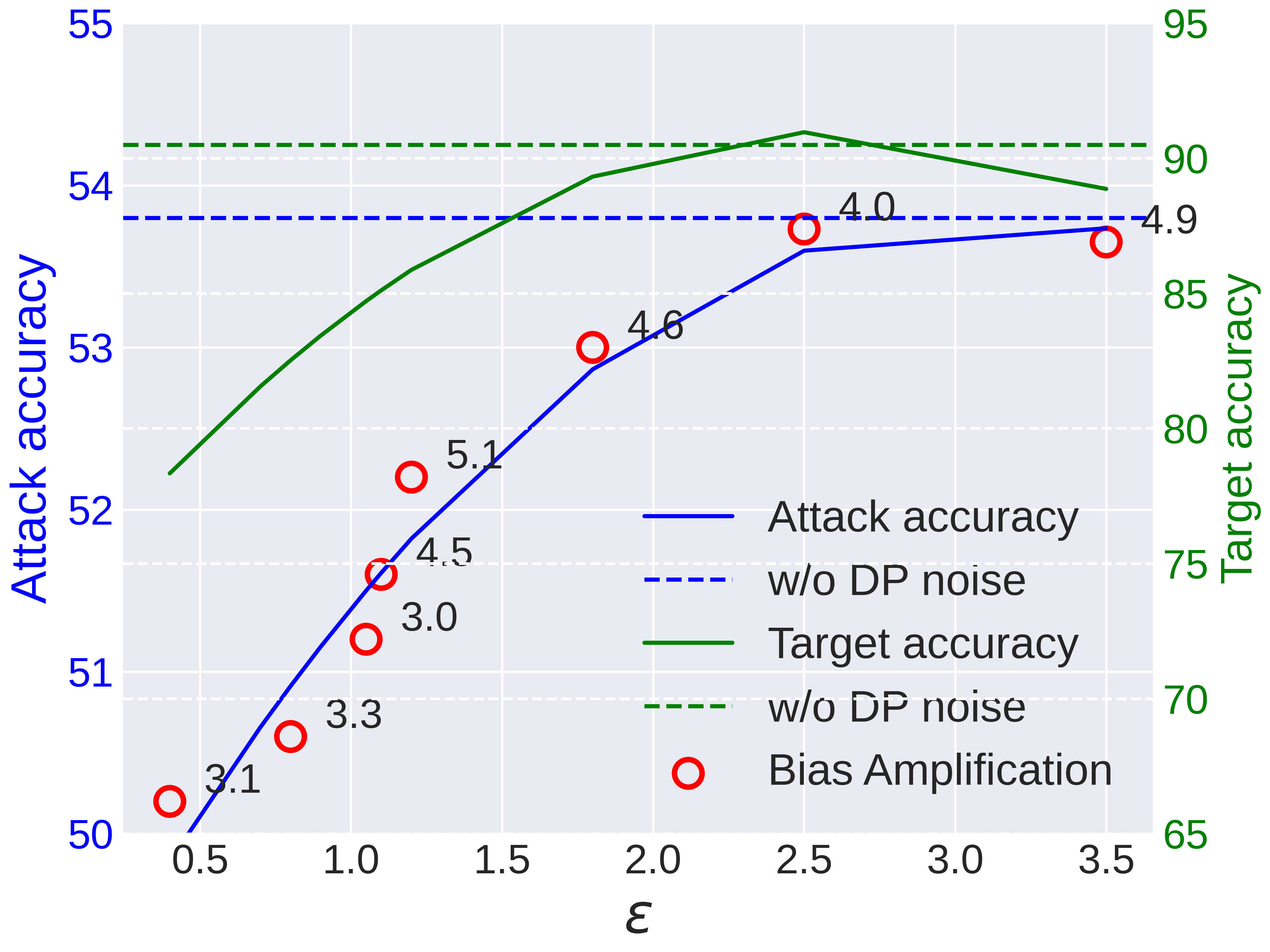}
    \caption{Attack results with DP-SGD with different values of $\epsilon$.}
    \label{fig:dp}
\end{figure}

\section{Attacks with varying skewed distributions}
We further consider varying skewed data distributions for the considered sensitive attribute. Specifically, we consider smiling classifications with the CelebA dataset considering \textit{gender} as the sensitive attribute and set imbalanced ratios between the majority and minority subgroups with different values ranging from $0.95$ to $0.75$. Table~\ref{table_dist} presents the results for target learning and the score-based attack results. All results indicate that fairness interventions tend to introduce some robustness to MIAs. However, with the proposed attack method, fair models can compromise model privacy with superior MIA results. The results are consistent with our findings.

\section{Results for mitigation}
To mitigate the privacy leakage of proposed methods, we adopt the Differential privacy (DP) technology for the defenses. We realize the method with Opacus\cite{opacus} library. We present the defense results with a fixed value of $\epsilon$ in the main paper. Here, we extend the evaluations and further compare the defense performance with different values of $\epsilon$ in Figure~\ref{fig:dp}. Given biased models, we enforce fairness interventions with DP noises. We then measure the accuracy of learning targets, fairness metric results (Bias Amplification), and attack performance. As shown in the figure, the DP noises will lead to decreased accuracy of learning targets. With smaller values of $\epsilon$, more noises are injected during the training, leading to inferior attack performance but lower prediction performance. In terms of fairness performance, with different values of $\epsilon$, all models can achieve fairness predictions compared with the results of biased models in Table~\ref{table:result_gender}. However, the fairness performance is inferior compared with fair-enforce models without DP noises. The figure indicates that DP-SGD can protect the privacy leakage of fair-enforced models while maintaining the performance of target predictions and fairness protections.

\end{document}